\documentclass[11pt]{article}

\usepackage{microtype}
\usepackage{graphicx}
\usepackage{subfigure}
\usepackage{booktabs} %

\usepackage[margin=2.9cm]{geometry}
\PassOptionsToPackage{linktocpage}{hyperref}
\usepackage[hyperindex,breaklinks,colorlinks,linkcolor=blue,citecolor=blue,bookmarks=false,pagebackref]{hyperref}
\title{Implicit Bias of Linear RNNs}
\author{Melikasadat Emami$^\dagger$, Mojtaba Sahraee-Ardakan$^{\dagger,\#}$, Parthe Pandit$^{\dagger,\#}$,\\[5pt] Sundeep Rangan$^\diamondsuit$, Alyson K. Fletcher$^{\dagger,\#}$\\[15pt]
\small$^\dagger$Department of Electrical and Computer Engineering, UCLA,\\
\small$^\#$Department of Statistics, UCLA, \\
\small$^\diamondsuit$Department of Electrical and Computer Engineering, NYU.}
\date{} 

\usepackage{algorithm,algorithmic}
\usepackage{verbatim}
\usepackage{amsmath}  
\usepackage{amssymb,amsthm,amsmath,amsfonts,mathtools}
\usepackage[inline]{enumitem}

\usepackage{mathtools}
\usepackage{tikz}
\usetikzlibrary{arrows}
\usepackage{amsthm}
\usepackage{bbm}
\usepackage{sidecap}
\usepackage{wrapfig}

\newcommand{\arr}{\rightarrow}
\newcommand{\Xmc}{\mathcal{X}}
\def\tr{\mathrm{tr}}

\newcommand{\Norm}{\mathcal{N}}
\newcommand{\Real}{\mathbb{R}}

\newcommand{\Exp}{\mathbb{E}}
\newcommand{\Prob}{\mathbb{P}}

\newcommand{\inv}{^{-1}}

\newcommand{\inner}[1]{\left< #1 \right>}

\newtheorem{theorem}{Theorem}[section]
\newtheorem{lemma}[theorem]{Lemma}
\newtheorem{proposition}[theorem]{Proposition}

\newenvironment{definition}[1][Definition]{\begin{trivlist}
\item[\hskip \labelsep {\bfseries #1}]}{\end{trivlist}}

\newcommand{\cov}{{\rm cov}}

\newcommand{\mcE}{\mathcal{E}}

\newcommand{\mc}[1]{\mathcal{#1}}
\newcommand{\wt}[1]{\widetilde{#1}}
\newcommand{\wh}[1]{\widehat{#1}}

\newcommand{\mbb}[1]{\mathbb{#1}}
\newcommand{\xbf}{\mathbf{x}}

\newcommand{\eqd}{\stackrel{d}{=}}

\def\beq{\begin{equation}}
\def\eeq{\end{equation}}
\def\beqstar{\begin{equation*}}
\def\eeqstar{\end{equation*}}

\newcommand{\ones}{\mathbf{1}}

\newcommand{\htil}{\widetilde{h}}
\newcommand{\zt}{\widetilde{z}}
\newcommand{\Zt}{\widetilde{Z}}
\newcommand{\yt}{\widetilde{y}}

\def\bm{\begin{matrix}}
\def\bbm{\begin{bmatrix}}
\def\ebm{\end{bmatrix}}

\newcommand{\norm}[1]{\|#1\|}
\newcommand{\Ab}{\mathbf{A}}
\newcommand{\Db}{\mathbf{D}}
\newcommand{\Ib}{\mathbf{I}}

\newcommand{\bb}{\mathbf{b}}
\newcommand{\xb}{\mathbf{x}}

\newcommand{\ub}{\mathbf{u}}
\newcommand{\yb}{\mathbf{y}}

\newcommand{\thb}{\mathbf{\theta}}
\newcommand{\rhob}{\mathbf{\rho}}
\DeclareMathOperator*{\I}{\mathbf{I}}

\newcommand{\tran}{^{\text{\sf T}}}

\def\PLeq{\stackrel{PL(2)}{=}}

\newcommand{\Mmc}{\mathcal{M}}

\newcommand{\Gbar}{\overline{G}}
\newcommand{\nlim}{ \lim_{n \rightarrow \infty}}
\newcommand{\suml}{\sum_{\ell = 1}^{L}}
\newcommand{\sumt}{\sum_{t' = 1}^{t}}
\newcommand{\sumtt}{\sum_{t'' = 1}^{t}}
\newcommand{\sumj}{\sum_{j = 1}^{t}}

\newcommand{\Rmc}{\mathcal{R}}

\newcommand{\Hmct}{\widetilde{\mathcal{H}}}
\newcommand{\ran}{{\rm ran}}
\newcommand{\dete}{{\rm det}}

\definecolor{ppink}{rgb}{0.858, 0.188, 0.478}

\newcommand{\At}{\widetilde{A}}

\newcommand{\Pt}{\widetilde{P}}

\newcommand{\Rt}{\widetilde{R}}

\newcommand{\rt}{\widetilde{r}}

\newcommand{\Ht}{\widetilde{H}}

\RequirePackage{bm}

\newcommand{\Tc}{\mathcal{T}}

\usepackage{etoolbox}
\newtoggle{conference}
\togglefalse{conference} %

\begin{document}

\onecolumn
\maketitle

\begin{abstract}

Contemporary wisdom based on empirical studies suggests that standard recurrent neural networks (RNNs) do not perform well on tasks requiring long-term memory. However, precise reasoning for this behavior is still unknown. This paper provides a rigorous explanation of this property in the special case of linear RNNs.  Although this work is limited to linear RNNs, even these systems have traditionally been difficult to analyze due to their non-linear parameterization. Using recently-developed kernel regime analysis, our main result shows that linear RNNs learned from random initializations are functionally equivalent to a certain weighted 1D-convolutional network.  Importantly, the weightings in the equivalent model cause an implicit bias to elements with smaller time lags in the convolution, and hence shorter memory.  The degree of this bias depends on the variance of the transition kernel matrix at initialization and is related to the classic exploding and vanishing gradients problem. The theory is validated in both synthetic and real data experiments.

\end{abstract}
\section{Introduction}

Over the past decade,  models based on neural networks have become commonplace in almost all machine learning applications. A key feature about this class of models is that, in principle, they do not require an explicit design of features but instead rely on an implicit learning of ``\textit{meaningful}'' representations of the data. This makes neural network models attractive from the point of view of machine perception where raw signals are inputs to the model. While neural networks have become ubiquitous in practical applications, several theoretical aspects of them largely remain a mystery. Of significant importance is the lack of understanding of the potential implicit biases that these representations bring to the predictions of the model.

This paper aims to understand the \textit{implicit bias} behaviour of Recurrent Neural Networks (RNN). Several machine learning tasks require dealing with sequential data with possibly varying lengths of input sequences. Some example tasks include automatic speech recognition, language translation, and image captioning, among others. In such tasks the simplest architecture is an RNN. Due to their simplicity, these models may be preferred over their complex descendants, Long Short-term Memory (LSTM) networks, in applications demanding interpretability.

A common critique of RNNs trained using gradient descent is their poor performance at tasks requiring long-term dependence \cite{bengio1994learning}. However, we lack a quantitative understanding of this phenomenon. In this paper, we provide a rigorous characterization for the implicit bias of RNNs towards short-term contexts. This characterization can suggest bias-correction strategies enabling the use of RNNs, instead of complex models such as LSTMs, for tasks that require long-term dependence. 

The use of RNN-based operators in convolutional networks is shown to significantly reduce the inference-time  memory for a wide range of tasks \cite{saha2020rnnpool}. Since the accuracy remains the same with the replacement of convolutional blocks with RNN based operators, the computational cost can be reduced dramatically in a lot of architectures.

To our knowledge this paper is the first that provides a characterization of the implicit bias in RNNs. Our analysis is based on two key observations. First, we show that linear RNNs are functionally equivalent to a 1D-convolutional model which is feed-forward in nature. Secondly, due to the Neural Tangent Kernel (NTK) regime based analysis \cite{jacot2018neural}, we are able to show that RNNs trained using gradient descent learn a subset of 1D-convolutional models --- those with short-term contexts.
Our result holds in a certain wide limit regime where the number of hidden units in RNN goes to infinity. 
We summarize the main contributions of this paper below.

\paragraph*{Main Contributions:}
\begin{itemize}
    \item We explicitly compute the NTK for a linear RNN.  This is challenging
    due to the weight sharing in RNNs which leads to statistical dependencies across time. We calculate this NTK using a conditioning technique as in \cite{bayati2011dynamics} to deal with the dependencies. This NTK is also calculated in \cite{alemohammad2020recurrent} using the Tensor program results of \cite{yang2019scaling}. 
    
    \item We show that the linear RNN NTK is equivalent to the NTK of a  scaled convolutional model with certain scaling coefficients. This means that in the wide limit regime (number of hidden units in RNN $\rightarrow \infty$), gradient descent training of a linear RNN with non-linear parameterization is identical to the training of an appropriately scaled convolutional model.
    \item The above results rigorously show that there is an implicit bias in using the non-linear
    parameterization associated with a linear RNN. In particular, training linear RNNs with non-linear parameterization using gradient descent is implicitly biased towards short memory.
    \item We demonstrate the bias-variance trade-off of linear RNNs in experiments on
    synthetic and real data.
\end{itemize}

\paragraph*{Prior Work}
The connection between kernel methods and infinite width neural networks was first introduced in \cite{Neal_1996}. Neural networks in the infinite width limit are equivalent to Gaussian processes at initialization and several papers have investigated the correspondence to kernel methods for a variety of architectures \cite{lee2018deep, novak2019bayesian, garriga-alonso2018deep, yang2019scaling, daniely2016toward, matthews2018gaussian}.  In particular, \cite{daniely2016toward} introduced a framework to link a reproducing kernel to the neural network and stochastic gradient descent was shown to learn any function in the corresponding RKHS if the network is sufficiently wide \cite{daniely2017sgd}.

A recent line of work has shown that gradient descent on over-parameterized networks can achieve zero training error with parameters very close to their initialization \cite{allen2018convergence, du2018gradient, du2018gradient2, li2018learning, zou2018stochastic}.
The analysis of the generalization error in this high dimensional regime led to exact characterizations of the test error for different architectures \cite{montanari2019generalization, hastie2019surprises, belkin2019reconciling, emami2020generalization, Barbier_2019, gerbelot2020asymptotic}.
In addition to convergence to a global minimum for an over-parameterized two-layer neural network, \cite{E_2020} also showed that the resulting functions are uniformly close to the ones found with the kernel regime. It was shown by \cite{jacot2018neural} that the behavior of an infinitely wide fully-connected neural network trained by gradient descent is characterized by the so-called Neural Tangent Kernel (NTK) which is essentially the linearization of the network around its initialization. The NTK was later extended for different architectures \cite{arora2019exact, yang2019scaling, yang2019wide, alemohammad2020recurrent}.

A different line of papers investigated the over-parameterized neural networks from the mean field viewpoint \cite{mei2018mean, wei2019regularization, ma2019comparative, dou2020training, sirignano2020mean, rotskoff2018neural}. For recurrent neural networks in particular, \cite{chen2018dynamical} has provided a theory for signal propagation in these networks which could predict their trainability. The authors also give a closed-form initialization to improve the conditioning of input-output Jacobian.

Trainablility of RNNs has also been improved by using orthogonal/unitary weight matrices \cite{arjovsky2016unitary, wisdom2016full, jing2017tunable, emami2019input}. It has been shown in \cite{emami2019input} that for RNNs with ReLU activations, there is no loss in the expressiveness of the model when imposing the unitary constraint.

Note that in an RNN the states are correlated due to weight sharing. Previous work such as \cite{chen2018dynamical}, has simplified the setting by assuming an independence over RNN weights (ignoring the correlation) to show that the pre-activations are Gaussian distributed. In this work, taking into account these dependencies, we use techniques used in \cite{bayati2011dynamics, rangan2019vector} to characterize the behaviour of RNNs at initialization. Similar techniques have also been explored in \cite{yang2019scaling}.

We should mention that learning the weight matrices of a linear RNN using data is essentially a system identification task. There is a large body of literature in control theory that consider the system identification problem and propose many different methods to find a system that matches the input-output behavior of a given system. These methods include the prediction error method (PEM), subspace methods, empirical transfer function estimate (ETFE), correlation method, spectral analysis method, and sequential Monte Carlo method to name a few. For a more comprehensive list of system identification methods and details see \cite{ljung1999system,ljung1994modeling,lennart1999system, katayama2006subspace,viberg1995subspace, soderstrom1989system,pintelon2012system, sjoberg1995nonlinear}.
Even though it would be interesting to see how different system identification methods can be incorporated into neural network training pipeline, the vast majority of works currently learn the weights directly by optimizing a loss function via gradient descent or its variants. As such, in this work we solely focus on training of linear RNNs using gradient descent.

\section{Linear RNN and Convolutional Models}  \label{sec:models}

\paragraph*{Linear RNNs}  We 
fix a time period $T$ and 
consider a linear RNN mapping 
an input sequence $\xb=(x_0,\ldots,x_{T-1})$ 
to an ouptut sequence $\yb=(y_0,\ldots,y_{T-1})$
via the updates
\begin{align} \label{eq:linrnn}
    h_{t} =\frac{1}{\sqrt{n}} Wh_{t-1} + Fx_t, \quad y_t = \frac{1}{\sqrt{n}}Ch_t, \quad
    t=0,\ldots,T-1,
\end{align}
with the initial condition $h_{-1} = \bm{0}$. 
We let $n_x$, $n_y$, and $n$ be the dimension at each time
of the input, $x_t$, output, $y_t$, and 
hidden state $h_t$ respectively.
Note that a bias term can be added for $h_t$ by extending $x_t$ and $F$.
We will let 
\begin{equation} \label{eq:frnn}
    \yb = f_{\rm RNN}(\xb,\theta_{\rm RNN}).
\end{equation}
denote the mapping \eqref{eq:linrnn} where $\theta_{\rm RNN}$
are the parameters
\begin{equation} \label{eq:theta_rnn}
    \theta_{\rm RNN} = (W,F,C).    
\end{equation}
The goal is to learn parameters $\theta_{\rm RNN}$ 
for the system from $N$ training data samples
$(\xb_i, \yb_i)$, $i=1,\ldots,N$.
In this case, each sample $(\xb_i, \yb_i)$ is a $T$-length input-output pair.

\paragraph{Wide System Limit}
We wish to understand learning of this system
in the \emph{wide-system limit} where the number of hidden units $n\rightarrow \infty$
while the dimensions
$n_x,n_y$ and number of time steps $T$ are fixed.
Since the parameterization of the RNN is non-linear, the initialization is critical.
For each $n$, we will assume that the parameters $W,F,C$ are initialized 
with i.i.d.\ components,
\begin{equation} \label{eq:thetaiid}
    W_{ij} \sim \Norm(0,\nu_W), \quad
    F_{ij} \sim N(0,\nu_F), \quad
    C_{ki} \sim N(0,\nu_C),
\end{equation}
for constants $\nu_W, \nu_F, \nu_C$.

\paragraph{Stability} 
In the initialization \eqref{eq:thetaiid},
$\nu_W$ is the variance of the components
of the kernel matrix 
$W$.  One critical aspect in selecting $\nu_W$
is the \emph{stability} of the system.
A standard result in linear systems theory (see, e.g.\ \cite{kailath1980linear}) is that the system
\eqref{eq:linrnn} is stable if and only if
$\frac{1}{\sqrt{n}}\lambda_{max}(W) < 1$
where $\lambda_{\rm max}(W)$ is the maximum absolute
eigenvalue of $W$ (i.e.\ the spectral radius).
Stable $W$ are generally necessary for linear RNNs:
Otherwise bounded inputs $x_t$ 
can result in outputs $y_t$ that grow unbounded with time
$t$.  Hence, training will be numerically unstable.
Now, a classic result in random matrix
theory \cite{bai1986limiting} is that, since the
entries of $W$ are i.i.d.\ Gaussian $\Norm(0,\nu_W)$,
\[
    \lim_{n \rightarrow \infty}\frac{1}{\sqrt{n}}
    \lambda_{\rm max}(W) = \nu_W
\]
almost surely.  Hence, for stability we need to select
$\nu_W < 1$.  As we will see below, it is this constraint
that will limit the ability of the linear RNN
to learn long-term memory.

\paragraph*{Scaled 1D Convolutional Equivalent Systems:}
Our main result will draw an equivalence between the learning of linear RNNs
and certain types of linear convolutional models.
Specifically, consider a \emph{linear convolutional model} of the form,
\beq  \label{eq:yconv}
    y_t = \sum_{j=0}^t L_j x_{t-j},
\eeq
where $L_j \in \Real^{n_y \times n_x}$ are the filter coefficient matrices.
In neural network terminology, the model \eqref{eq:yconv} is a simply a 
linear 1D convolutional network with $n_x$ input channels, $n_y$ output channels
and $T$-wide kernels. 

Both the linear RNN model \eqref{eq:linrnn} and the 1D convolutional model \eqref{eq:yconv}
 define linear mappings of
$T$-length input sequences $\xb$ to $T$-length output sequences $\yb$.
To state our equivalence result between these models, we need to introduce
a certain \emph{scaled parametrization}:  
Fix a set of scaling factors $\rhob = (\rho_0,\ldots,\rho_{T-1})$
where $\rho_j > 0$ for all $j$.
Define the parameters
\begin{equation} \label{eq:theta_conv}
    \theta_{\rm conv} = (\theta_0,\ldots,\theta_{T-1}),
\end{equation}
where $\theta_j \in \Real^{n_y \times n_x}$.  
Given any $\theta$, let the impulse response coefficients be
\begin{equation}  \label{eq:conv:lwt}
    L_j = \sqrt{\rho_j} \theta_j.
\end{equation}
As we will see below, the effect of the weighting is to favor certain
coefficients $L_j$ over others during training:  
For coefficients $j$ where $\rho_j$
is large, the fitting will tend to select $L_j$ large if needed.
This scaling will be fundamental in understanding implicit bias.

Now, given a set of weights $\rho=(\rho_0,\ldots,\rho_{T-1})$, let
\begin{equation} \label{eq:fconv}
    \yb = f_{\rm conv}(\xb,\theta_{\rm conv}) := \left\{\sum_{j=0}^t \sqrt{\rho_j} \theta_j x_{t-j}\right\}_{t=0}^{T-1}, 
\end{equation}
denote the mapping of the  
input $\xb=(x_0,\ldots,x_{T-1})$ through the convolutional filter
$\theta_{\rm conv}$ with filter coefficients $\rho$
to produce the output sequence $\yb=(y_0,\ldots,y_{T-1})$.

It is well-know that 
the RNN and convolutional models define the same total set of
input-output mappings as given by the following standard result:

\begin{proposition}\label{prop:lin-RNN_equivalence}
Consider the linear RNN model \eqref{eq:frnn} and 
the 1D convolutional model \eqref{eq:fconv}.
\begin{enumerate}[label=\emph{(\alph*)}]
    \item Given a linear RNN parameters $\theta_{\rm RNN}=(W,F,C)$, 
    a 1D convolutional filter with coefficient matrices
\begin{equation} \label{eq:linrnn_impulse}
    L_j = \frac{1}{n^{\frac{j+1}{2}}}CW^jF, \quad j = 0,\ldots,T-1,
\end{equation}
    will have an identical input-output mapping.  That is, there exists
    parameters $\theta_{\rm conv}$ such that
    \begin{equation}
        f_{\rm RNN}(x,\theta_{\rm RNN}) = 
        f_{\rm conv}(x,\theta_{\rm conv}),
    \end{equation}
    for all inputs $x$.
    \item Conversely, given any $T$ filter coefficients $\{L_j\}_{j=0}^{T-1}$, 
    there exists RNN model with $n \leq T n_x n_y$ hidden states such that the RNN and 1D convolutional model have identical input-output mappings
    over $T$-length sequences.
\end{enumerate}  
\end{proposition}
\begin{proof}
These are standard results from linear systems theory \cite{kailath1980linear}.
In the linear systems theory, the coefficients $L_i$ are together called the \emph{matrix impulse response}.  Part (b) follows by finding $C$, $W$ and
$F$ to match the equations \eqref{eq:linrnn_impulse}.  
Details are given in the Appendix \ref{sec:equiv_proof}.
\end{proof}

\paragraph*{Linear and Non-Linear Parametrizations:}
Proposition \ref{prop:lin-RNN_equivalence} shows that 
linear RNNs with sufficient width can represent the same
input-output mappings as any linear convolution system.  
The difference between the
models is in the parameterizations.  The output of the convolutional model is linear
in the parameters $\theta_{\rm conv}$ 
whereas it is non-linear in $\theta_{\rm RNN}$. 
As we will see below, the non-linear parameterization 
of the RNN results in certain implicit biases.

\section{NTKs of Linear RNNs and Scaled Convolutional Models} \label{sec:NTK_background}

\subsection{Neural Tangent Kernel Background} 

To state our first set of results, we briefly review the 
neural tangent kernel (NTK) theory from \cite{jacot2018neural, arora2019exact}.  
The main definitions and results
we need are as follows:
Consider the problem
of learning a (possibly non-linear) model of the form
\begin{equation}  \label{eq:fnonlin}
    \wh{y} = f(x,\theta),
\end{equation}
where $x \in \Real^{m_x}$ is an input, $f(\cdot)$ is a model function differentiable with respect to 
 parameters $\theta$,
and $\wh{y}$ is some prediction of an output $y \in \Real^{m_y}$.
The problem is to learn the parameters $\theta$ from training data
$\{x_i,y_i\}_{i=1}^N$.
For sequence problems, we use the convention 
that each $x_i$ and $y_i$ represent one entire input-output sequence pair.  Hence, the dimensions
will be $m_x = n_xT$ and $m_y = n_yT$.

Now, given the training data $\{x_i,y_i\}_{i=1}^N$ and 
an initial parameter estimate $\theta^0$, the 
\emph{neural tangent kernel (NTK) model} is the linear model
\begin{align}  
    \wh{y} &= f^{\rm lin}(x,\alpha) 
    := f(x,\theta^0) + \sum_{j=1}^N K(x_i,x_j)\alpha_j, \label{eq:flin}
\end{align}
where $K(x,x')$ is the so-called NTK,
\begin{align}\label{eq:fc_ntk}
    [K(x,x')]_{ij} &:=  \inner{\frac{\partial f_i(x,\theta^0)}{\partial \theta},\frac{\partial f_j(x',\theta^0)}{\partial \theta}}.
\end{align}
and $\alpha$ is a vector of dual coefficients,
\begin{equation} \label{eq:alpha}
    \alpha = (\alpha_1,\ldots,\alpha_N), \quad \alpha_j\in \Real^{m_x}.
\end{equation}
Note that $K(x,x')$ depends implicitly on $\theta^0$.
Also, for a fixed initial condition, $\theta^0$, 
the model \eqref{eq:flin} is linear in the parameters $\alpha$,
and hence potentially easier to analyze than the original 
non-linear model \eqref{eq:fnonlin}.
The key result in NTKs is that, for certain
wide neural networks with random initializations, 
(full-batch) gradient descent training of the non-linear and linear models are 
\emph{asymptotically identical}.  For example, the results in 
\cite{lee2019wide} and \cite{alemohammad2020recurrent} provide the following proposition:

\begin{proposition}  \label{prop:ntk_conv}
Suppose that $f_n(x,\theta)$ is a sequence of 
recurrent neural networks with $n$ hidden states and non-linear activation function $\sigma(\cdot)$.
Let $\{(x_i,y_i)\}_{i=1}^N$ be some fixed training data contained in a compact set.
Let $\wh{\theta}^0_n$ denote a random initial condition generated as \eqref{eq:thetaiid} and let $\wh{\theta}^{\ell}_n$ denote the 
parameter estimate after $\ell$ steps of (full-batch) gradient descent with some learning 
rate $\eta$.  Let $K_n(x,x')$ denote the NTK of the RNN 
and $f_n^{\rm lin}(x,\alpha)$ denote
the corresponding linear NTK model \eqref{eq:flin}.  Let 
$\wh{\alpha}^{\ell}_n$ denote the parameter estimate obtained with GD with the same learning rate.
We further assume that the non-linear activation $\sigma$ satisfies
\begin{equation*}
    |\sigma(0)|,~~ \norm{\sigma'}_\infty, ~~\sup_{x\neq x'} |\sigma(x)-\sigma(x')|/|x-x'| < \infty. 
\end{equation*}
Then, for all $x$ and $x'$,
\begin{equation}
    \lim_n K_n(x,x') = K(x,x') \mbox{ a.s.}
\end{equation}
for some deterministic positive semi-definite matrix $K(x,x')$.
Moreover, if $\lambda_{\rm min}(K)>0$, then 
for sufficiently small learning rate $\eta$ and any new sample $x$,
\begin{equation}
    \lim_{n \rightarrow \infty} \sup_{\ell \geq 0} 
    \| f_n(x,\wh{\theta}^\ell_n) -
    f_n^{\rm lin}(x,\wh{\alpha}^\ell_n) \| = 0,
\end{equation}
where the convergence is in probability.
\end{proposition}

The consequence of this result is that the behavior of
certain infinitely-wide neural networks on new samples $x$
is identical to the behavior of the linearized network around its initialization. This essentially means that as $n \rightarrow \infty$, the learning dynamics for the original and the linearized networks match during training.

\subsection{NTK for the Convolutional Model}
Having defined the NTK, we first compute the NTK of the scaled convolutional model \eqref{eq:fconv}.

\begin{theorem} \label{thm:lin_param}
Fix a time  period $T$ and consider the convolutional model \eqref{eq:fconv} for a given set 
of scale factors $\mathbf{\rho} = (\rho_0,\ldots,\rho_{T-1})$.
Then, for any initial condition, and any two input sequences $\xb$ and $\xb'$, 
the NTK for this model is,
\beq \label{eq:conv_ntk}
    K(\xb,\xb') = \Tc(\xb)\tran D(\rho)\Tc(\xb') \otimes I_{n_y},
\eeq
where $\Tc(\xb)$ is the Toeplitz matrix,
\beq \label{eq:Txdef}
    \Tc(\xb) := \left[ \begin{array}{cccc}
         x_0 & x_1 & \cdots & x_{T-1} \\
          0  & x_0 & \cdots & x_{T-2} \\
         \vdots & \vdots & \ddots & \vdots  \\
         0 & 0 & \cdots & x_0 
    \end{array}\right] \in \Real^{Tn_x \times T}.
\eeq
and $D(\rho)$ is the diagonal matrix,
\beq \label{eq:Dalpha}
    D(\rho) := \mathrm{diag}(\rho_0 I_{n_x}, \cdots, \rho_{T-1} I_{n_x}).
\eeq
\end{theorem}
\begin{proof}
 See Appendix \ref{app:proof_lin_param} for proof.  
\end{proof}

\subsection{NTK for the RNN Parametrization}
We now compare the NTK of the scaled 
convolutional model to the NTK for the RNN parameterization.  

\begin{theorem} \label{thm:linrnn}
Fix a time  period $T$ and consider the RNN model \eqref{eq:linrnn}
mapping an input sequence $\xb=(x_0,\ldots,x_{T-1})$
to an ouptut sequence $\yb=(y_0,\ldots,y_{T-1})$
with the parameters  $(W,F,C)$.  Assume the parameters are initialized as
\eqref{eq:thetaiid} for some constants $\nu_W,\nu_F,\nu_C > 0$.
In the limit as the number of hidden states $n\rightarrow \infty$:
\begin{enumerate}[label=(\emph{\alph*})]
    \item The impulse response coefficients $L_j$ in \eqref{eq:linrnn_impulse}
    converge in distribution to independent Gaussians, where 
    the components of $L_j$ are i.i.d. $\Norm(0,\nu_C\nu_F \nu_W^j)$.
    
    \item Given any input sequences $\xb$ and $\xb'$, the NTK converges almost
    surely to the deterministic limit:
\beq \label{eq:lin_rnn_ntk}
    K(\xb,\xb') = \Tc(\xb)\tran D(\rho)\Tc(\xb') \otimes I_{n_y},
\eeq
where $\Tc(\xb)$ and $D(\rho)$ are given in \eqref{eq:Txdef}
and \eqref{eq:Dalpha} and 
\beq \label{eq:alphas}
    \rho_j = \nu_C(j\nu_F\nu_W^{j-1} + \nu_W^j) + \nu_F \nu_W^j.
\eeq
\end{enumerate}
\end{theorem}
\begin{proof}
 See Appendix \ref{app:linrnn_proof} for the proof.
\end{proof}

Comparing Theorems~\ref{thm:lin_param} and \ref{thm:linrnn}, we see that 
the NTK for linear RNN is identical to that of an scaled convolution model
when the scaling are chosen as \eqref{eq:alphas}.
From Proposition~\ref{prop:ntk_conv},
we see that, in the wide limit regime where $n \rightarrow \infty$, gradient descent training of the linear RNN
with the nonlinear parametrization $\theta_{\rm RNN}=(W,F,C)$ in \eqref{eq:theta_rnn} is 
identical to the training of the convolutional model
\eqref{eq:yconv} where the linear parameters $L_j$
are initialized as i.i.d.\ Gaussians and then 
trained with certain scaling factors \eqref{eq:alphas}.

Moreover, the scaling factors have a geometric decay.
Recall from Section \ref{sec:models} that, for stability
of the linear RNN we require that $\nu_W < 1$.
When $\nu_W < 1$ and $j > 1$,
the scaling factors \eqref{eq:alphas} can be bounded as
\begin{equation} \label{eq:rhodecay}
    \rho_j \leq \rho_{\rm max} \nu_W^{j-1}, \quad \rho_{\rm max} := \nu_C(T\nu_F + 1) + \nu_F.
\end{equation}
Consequently, the scale factors decay geometrically with $\nu_W^{j-1}$.  This implies that, in the training
of the scaled convolutional model, the coefficients
with higher delay $j > 1$ will be given lower
weight.

\section{Implicit Bias of Linear RNNs}

An importance consequence of the geometric
decay of the scaling factors $\rho_j$ in 
\eqref{eq:rhodecay} is the \emph{implicit bias}
of GD training of linear RNNs towards
networks with short-term memory.
To state this precisely,
fix input and output training data $(\xb_i,\yb_i)$, $i=1,\ldots,N$.
For each $n$, consider the RNN model 
\eqref{eq:frnn} with $n$ hidden states and parameters $\theta_{\rm RNN}$ in \eqref{eq:theta_rnn}.
Assume the parameters are initialized as $\theta^0_{\rm RNN}=(W^0,F^0,C^0)$
in \eqref{eq:thetaiid} for some $\nu_W,\nu_F,\nu_C > 0$.
Let 
\[
    \theta_{\rm RNN}^{\ell} = (W^\ell, F^\ell, C^\ell),
\]
denote the parameter
after $\ell$ steps of (full batch) gradient descent with some
learning rate $\eta$.  
Let $L_{\rm RNN}^{\ell}$ be the resulting impulse response coefficients
\eqref{eq:linrnn_impulse},
\begin{equation} \label{eq:rnn_imp_gd}
    L_{{\rm RNN},j}^\ell = n^{-(j+1)/2} C^\ell(W^\ell)^j 
    F^\ell, 
    \quad j=0,\ldots,T-1.
\end{equation}
We then have the following bound.

\begin{theorem}  \label{thm:implicit_bias}
Under the above assumptions, the norm of the impulse response coefficients
of the RNN at the initial iteration $\ell = 0$ are given by
\begin{equation} \label{eq:Lrnn_norm_init}
        \lim_{n \rightarrow \infty} 
    \Exp\| L_{{\rm RNN}, j}^0 \|^2_F = n_xn_y \nu_C\nu_F \nu_W^j.
\end{equation}
Also, there exists constants $B_1$ and $B_2$ such that if the learning rate 
satisfies $\eta < B_1$, then for all iterations $\ell$
\begin{equation} \label{eq:Lrnn_norm_iter}
    \limsup_{n \rightarrow \infty} 
    \| L_{{\rm RNN}, j}^\ell - L_{{\rm RNN}, j}^0 \|_F \leq B_2 \rho_j \eta \ell,
\end{equation}
where the convergence is in probability.  
Moreover, the constants $B_1$ and $B_2$ can be selected independent of $\nu_W$.
\end{theorem}
\begin{proof}
See Appendix \ref{app:implicit_bias_proof} for proof.
\end{proof}

To understand the significance of the theorem,
observe that in 
the convolutional model \eqref{eq:yconv},
$L_j$ relates the input time samples $x_{t-j}$ to the output
$y_t$.  The coefficient $L_j$ thus describes the
influence of the inputs samples on the output samples
$j$ time steps later.  
Combining \eqref{eq:Lrnn_norm_init}, \eqref{eq:Lrnn_norm_iter}, and \eqref{eq:rhodecay}, 
we see that these coefficients decay as
\[
    L_{{\rm RNN}, j}^\ell = O(\nu_W^{j/2} + \ell \nu_W^j).
\]
Also, as discussed in Section~\ref{sec:models}, we need $\nu_W < 1$
for stability.  
Hence, the magnitude of these  coefficients decay geometrically
with $\nu_W^{j-1}$.  Therefore, 
for a fixed number of 
training steps, the effect of the input on the output at a lag of $j$
would be exponentially small in $j$.  
In this sense, training linear RNNs with the non-linear
parameterization $\theta_{\rm RNN}=(W,F,C)$
is implicitly biased to short memory.

It is useful to compare the performance of an unscaled convolutional model with the linear RNN. 
The convolutional model can fit any linear time-invariant system with an arbitrary delay.
We have seen in Proposition~\ref{prop:lin-RNN_equivalence} that,
in principle, the linear RNN can also fit any such system with a sufficient number of hidden states.  
However, the above theorem shows that unless the number of gradient steps grows exponentially
with the desired delay, the parameterization of the linear RNN will strongly bias the solutions
to systems with short memory.  This restriction will create bias error on systems that have
long-term memory.  On the other hand, due to the implicit constraint of the linear RNN, 
the parameterization will reduce the variance error.

\section{Numerical Experiments}
We validate our theoretical results on a number of synthetic and real data experiments.
\subsection{Synthetic data}
In section \ref{sec:NTK_background}, we showed that the NTK for a linear RNN given in \eqref{eq:linrnn} with parameters $\theta_{\rm RNN}$ in \eqref{eq:theta_rnn} is equivalent to the NTK for its convolutional parameterization with parameters $\theta_{\rm conv}$. In order to validate this, we compared the training dynamics of a linear RNN, eq \eqref{eq:linrnn}, with a large number of hidden states ($n$) and a scaled convolutional model, eq \eqref{eq:fconv} with scale coefficients $\rho_j$ defined in \eqref{eq:alphas}. 
Our theory indicates that using gradient descent with the same learning rate, the dynamics of both models are identical during training, if they are initialized properly initialized. 

We generated data from a synthetic (teacher) RNN with random parameters. For the data generation system we used a linear RNN with 4 hidden units and $n_x=n_y=1$. Matrices $W$, $F$, and $C$ are generated as i.i.d random Gaussian with $\nu_W=0.3$ and $\nu_F = \nu_C=1$. We added noise to the output of this system such that the signal-to-noise ratio (SNR) is 20 dB. We generated 50 training sequences and 50 test sequences in total and each sequence has $T=10$ time steps.

\begin{figure}
\centering
\includegraphics[scale=0.5]{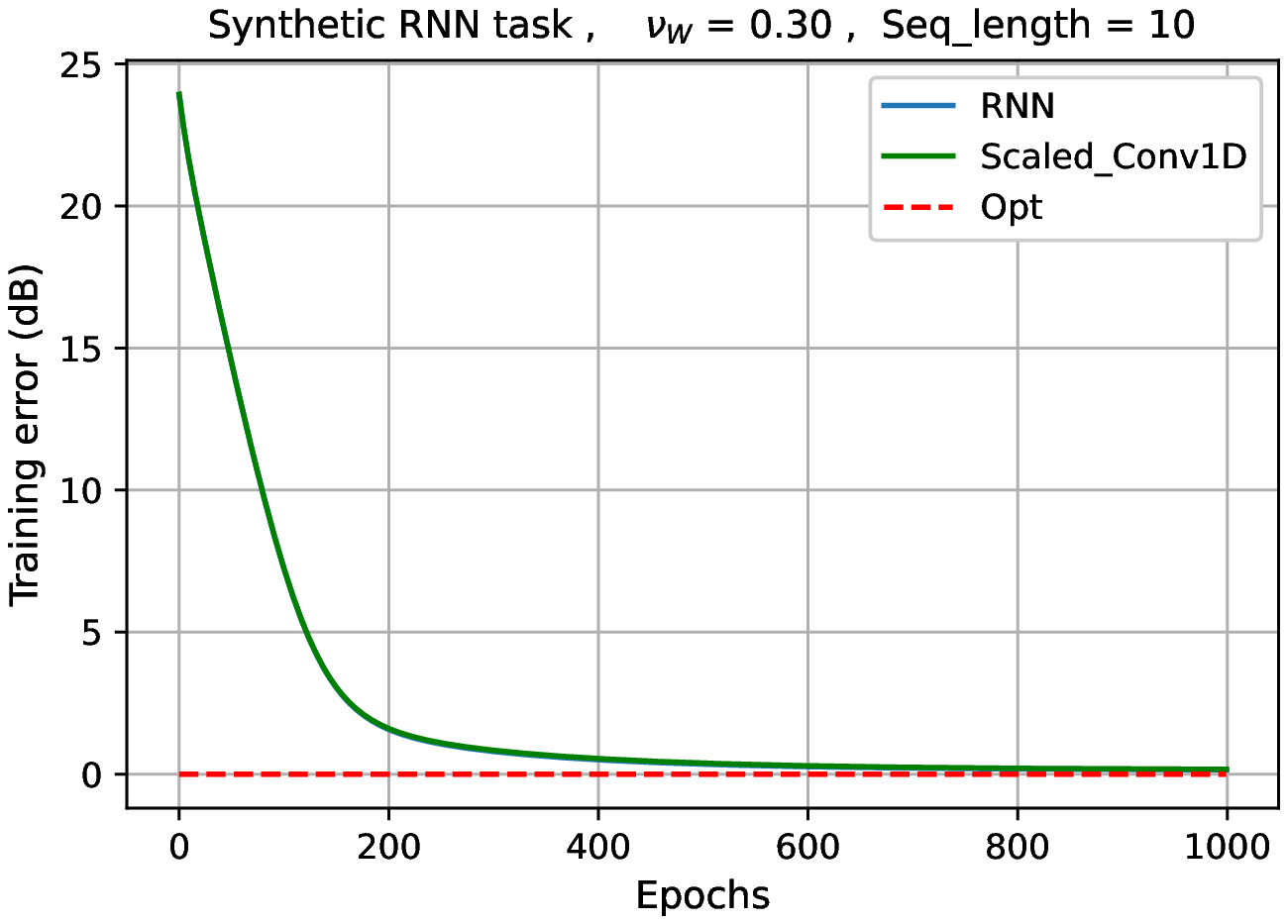}
\includegraphics[scale=0.5]{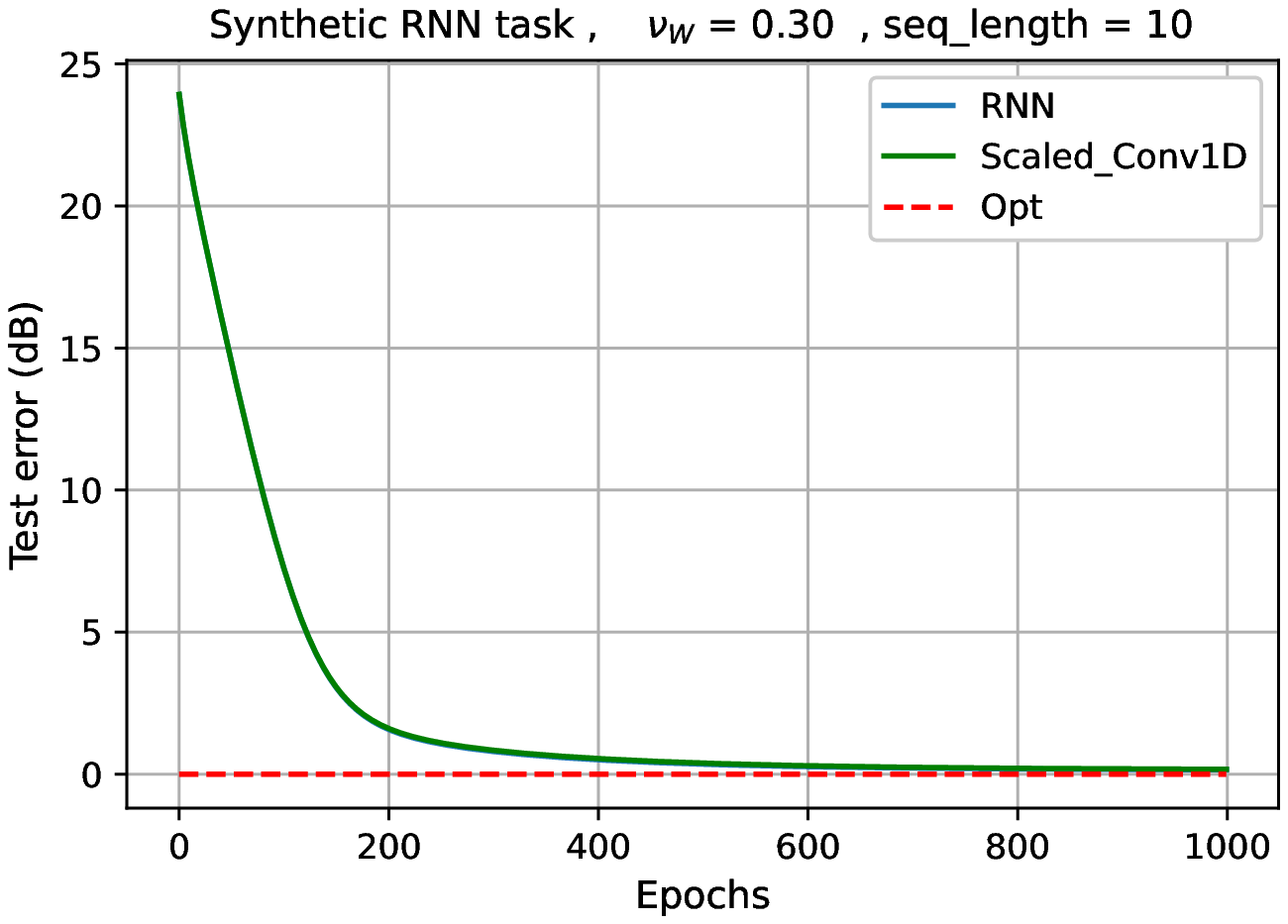}
\caption{Dynamics of an RNN and its equivalent scaled Conv-1D in learning a synthetic task. the data is generated from a synthetic RNN with $n_x=n_y=1$ and $nh=4$. Noise is added to the output with SNR$=20$ dB The sequence length T=10. Training and test samples are $N_{\rm tr} = N_{\rm ts} =50$. Full batch gradient descent is used with $ lr=10^{-4}$. As you see the dynamics of these models perfectly match.}
\label{fig:synth_tr}
\end{figure}

\begin{figure}
\centering
\includegraphics[scale=0.5]{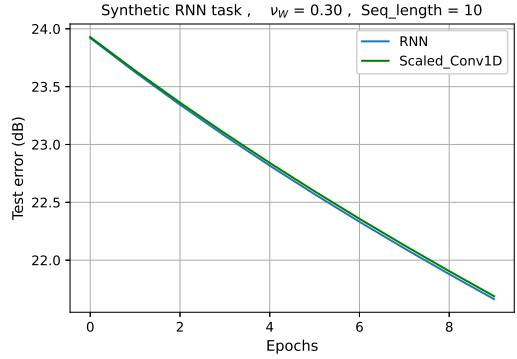}
\caption{The first 10 epochs in Fig.~\ref{fig:synth_tr}. Note that to be theoretically accurate, we need $lr \rightarrow 0$ to be in the kernel regime. }
\label{fig:synth_small}
\end{figure} 
      
\begin{figure}
\centering
\includegraphics[scale=0.5]{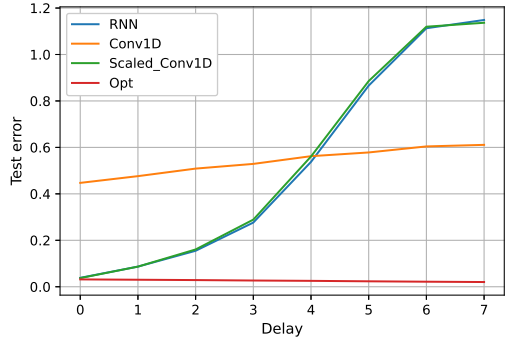}
\caption{Test performance with respect to delay. For this task we have $n_h = 1000$, $N_{\rm tr} = N_{\rm ts} = 10$, $n_x = 15$, $n_y=1$, and $T=20$. The delay is added manually by shifting i.e. $y_t = x_t - delay$ and the output SNR = 20 dB.}
\label{fig:delay}
\end{figure}

Given the training and test data, we train (i) a (student) linear RNN with $n = 1000$ hidden units, $\nu_W=0.3$, and $\nu_F = \nu_C=1$; and (ii) a 1D-convolutional model with scale coefficients $\rho_j$ calculated in \eqref{eq:alphas} using $\nu_W=0.3$, $\nu_F = \nu_C=1$. We used mean-squared error as the loss function for both models and applied full-batch gradient descent with learning rate $lr = 10^{-4}$.
Fig.~\ref{fig:synth_tr} shows the identical dynamics of training for both models. Fig.~\ref{fig:synth_small} shows a zoomed-in version of the dynamics for training error. Note that the theoretical convergence requires the $lr \rightarrow 0$ and $n \rightarrow \infty$ so that both models operate in the kernel regime.

To evaluate the performance of these models for a task with long-term dependencies, we created a dataset where we manually added different delay steps to the output of a true linear RNN system i.e. $y_t = x_t - delay$. We have chosen longer ($T=20$) true sequences for this task. We then learned this data using the aforementioned linear RNN and scaled 1D convolutional models. We also trained an unscaled 1D convolutional model with this data to compare performances. With unscaled convolutional model, we exactly learn the impulse response coefficients $L_j$ defined in \eqref{eq:yconv} during training. 

Fig.~\ref{fig:delay} shows the test error with respect to delay steps for all three models. Observe that the performance of the scaled convolutional and the linear RNN models match during training. Due to the bias of these models against the delay, the test error increases as we increase the delay steps in our system. On the other hand, the performance of the unscaled convolutional model stays almost the same with increasing delay, slightly changing at larger delays as there is less data to track.

As we increase the delay steps, the test error for these two models increases, and it is due to the fact that these models are biased against the delay.

\subsection{Real data} \label{sec:real_data}
We also validated our theory using spikes rate data from the macaque primary somatosensory cortex (S1) \cite{benjamin2018modern}. Somatosensory cortex is a part of the brain responsible for receiving sensations of touch, pain, etc from the entire body.
The data is recorded during a two-dimensional reaching task. In this task, a macaque was rewarded for positioning a cursor on a series of randomly generated targets on the screen using a handle. The data is from a single recording of 51 minutes and includes 52 neurons. The mean and median firing rates are $9.3$ and $6.3$ spikes/sec.%
 Similar to the previous  experiments, we also trained an unscaled 1D convolutional model with this data and compared the performances with the linear RNN and the scaled convolutional models. 
 
We compared the performances on two sets of experiments. We first used only the 4.5 minutes of the total recorded data. The purpose of this experiment is to compare the performances in limited data circumstances. With this limited data, we expect the scaled convolutional model (and thus the RNN) to perform better than the unscaled model due to the implicit bias of the towards short-term memory and the fact that the effective number of parameters is smaller in the scaled model which leads to a smaller variance. In the second experiment, we trained our models using all the available data ($\approx 51$ mins). In this case, the scaled model (and the RNN) performs worse because of the increased bias error.  
In our experiments setup, the linear RNN has $n=1000$ hidden states and the sequence length $T=15$. Also, $\nu_W = 0.3$ and $\nu_F=\nu_C=1$. 

\begin{figure}
\begin{tabular}{c|c }
 (a) Limited data  & (b) Entire data \\ 
 \includegraphics[scale=0.5]{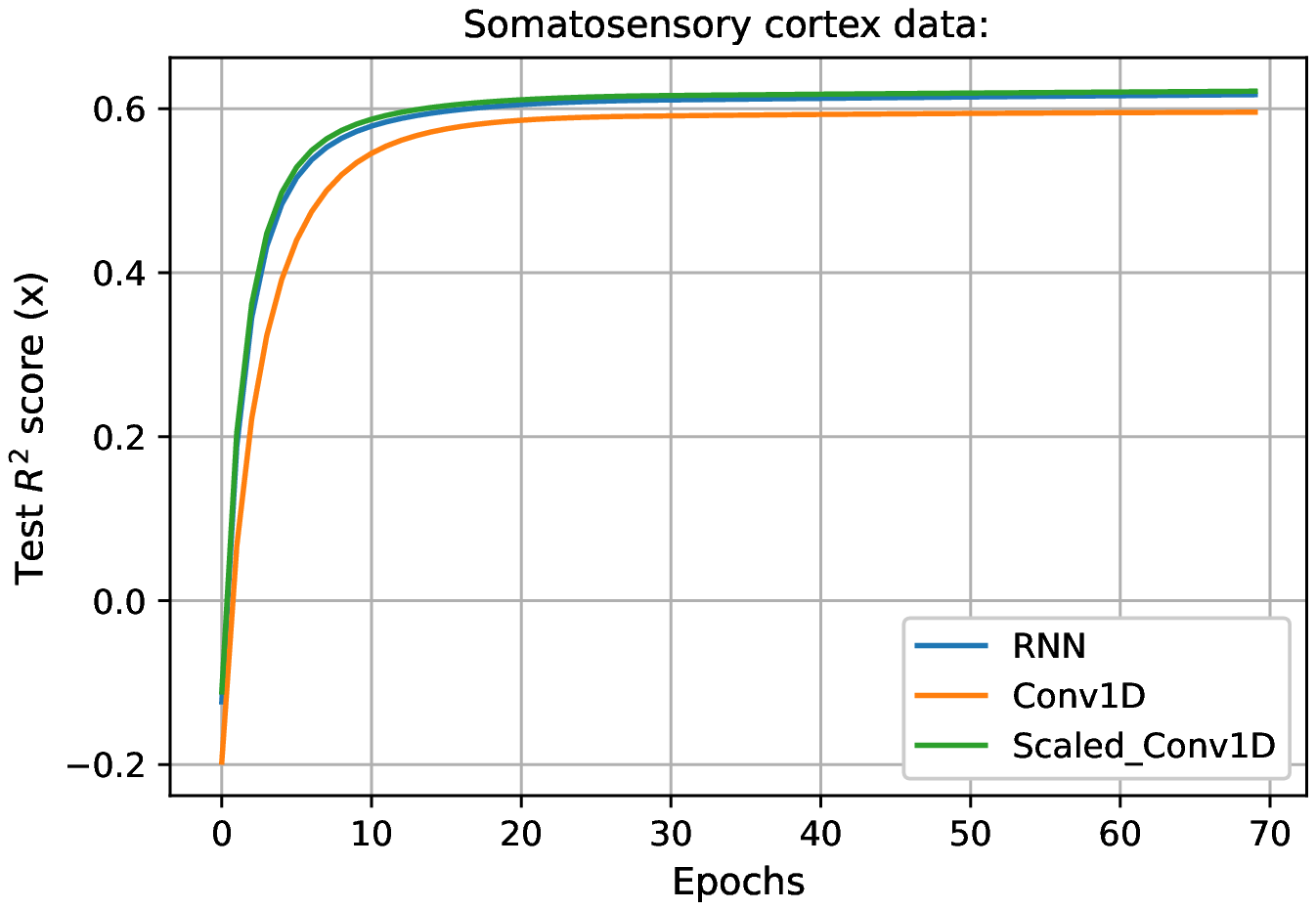}
 & \includegraphics[scale=0.5]{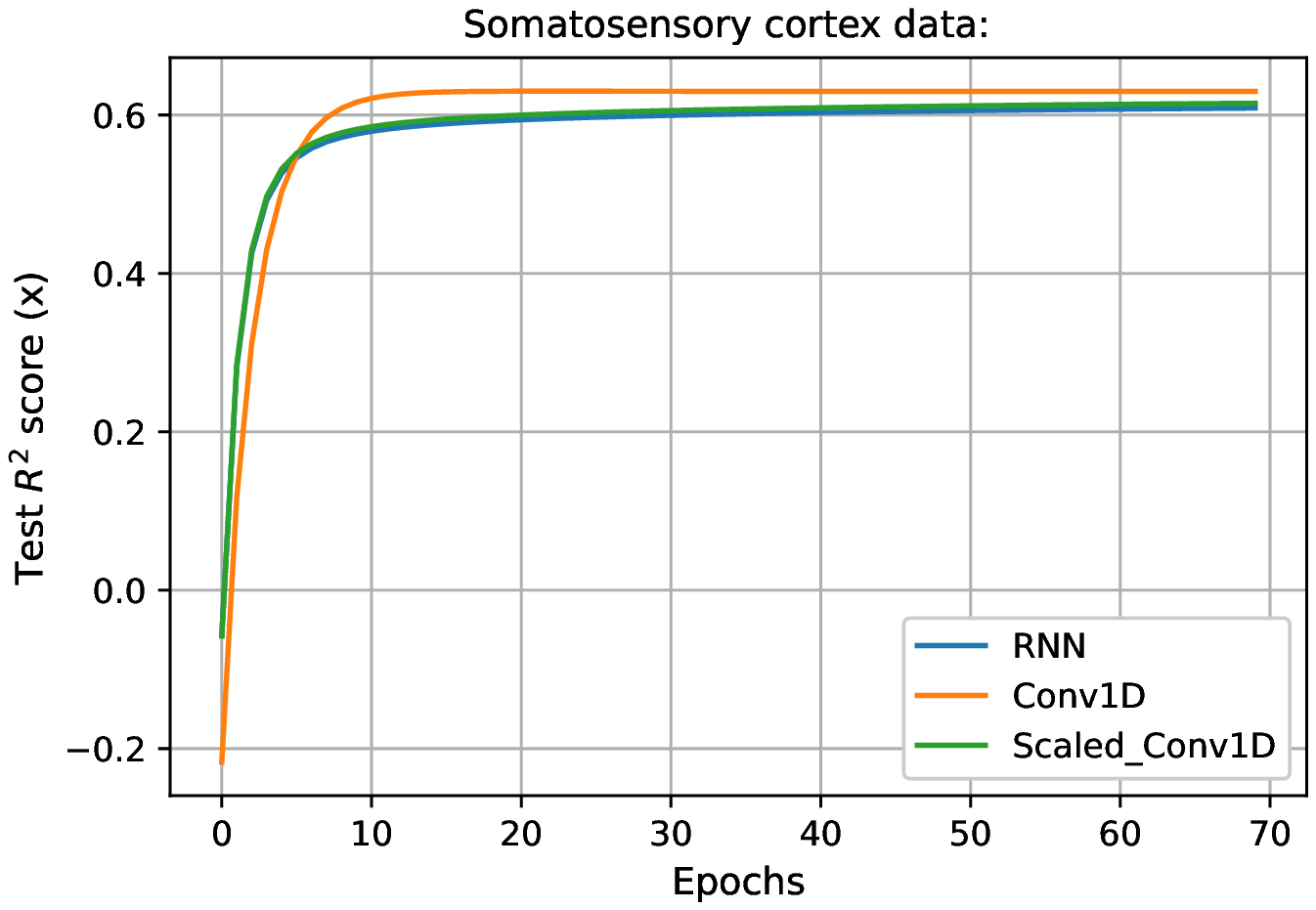}\\
 \includegraphics[scale=0.5]{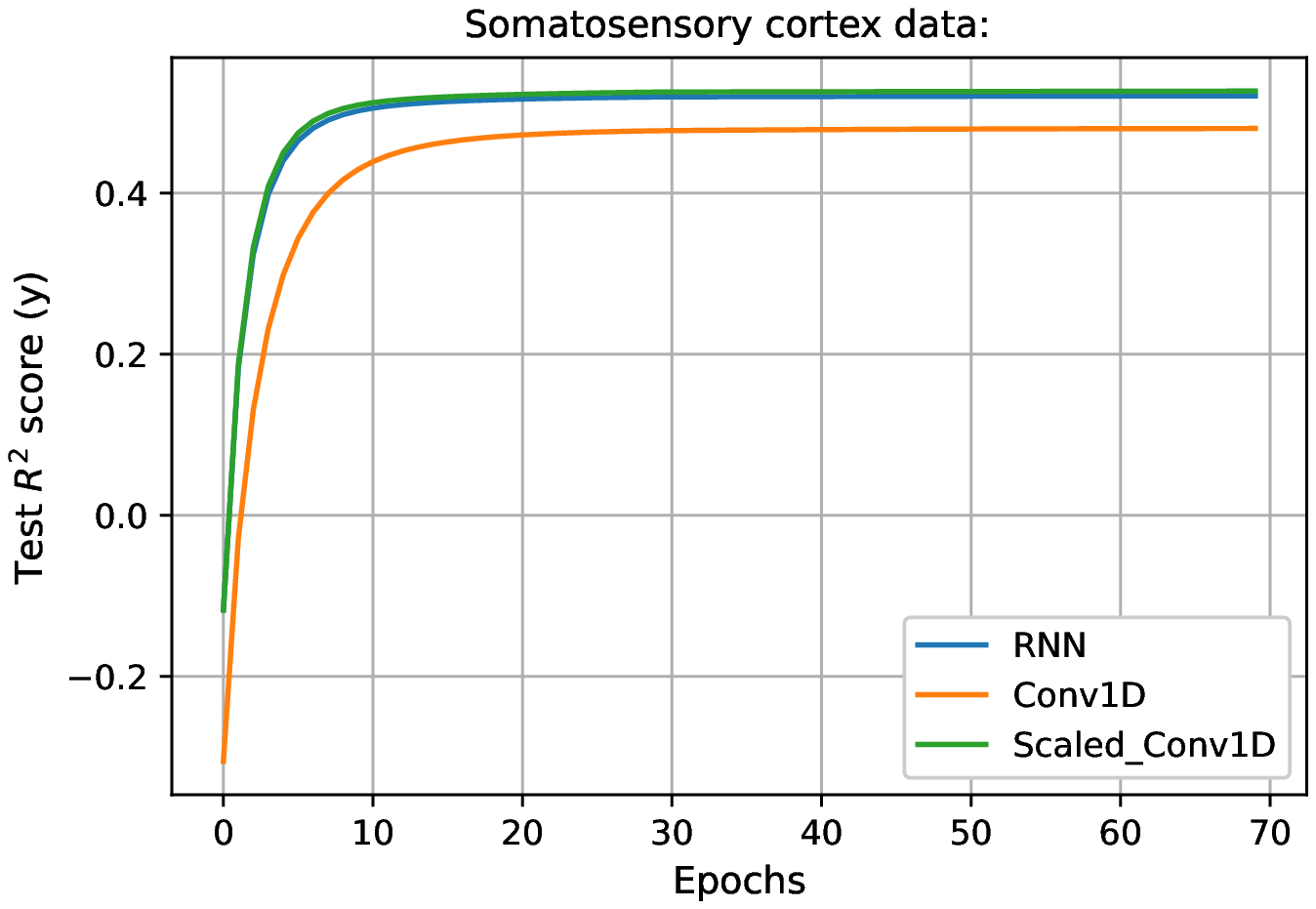}  &
\includegraphics[scale=0.5]{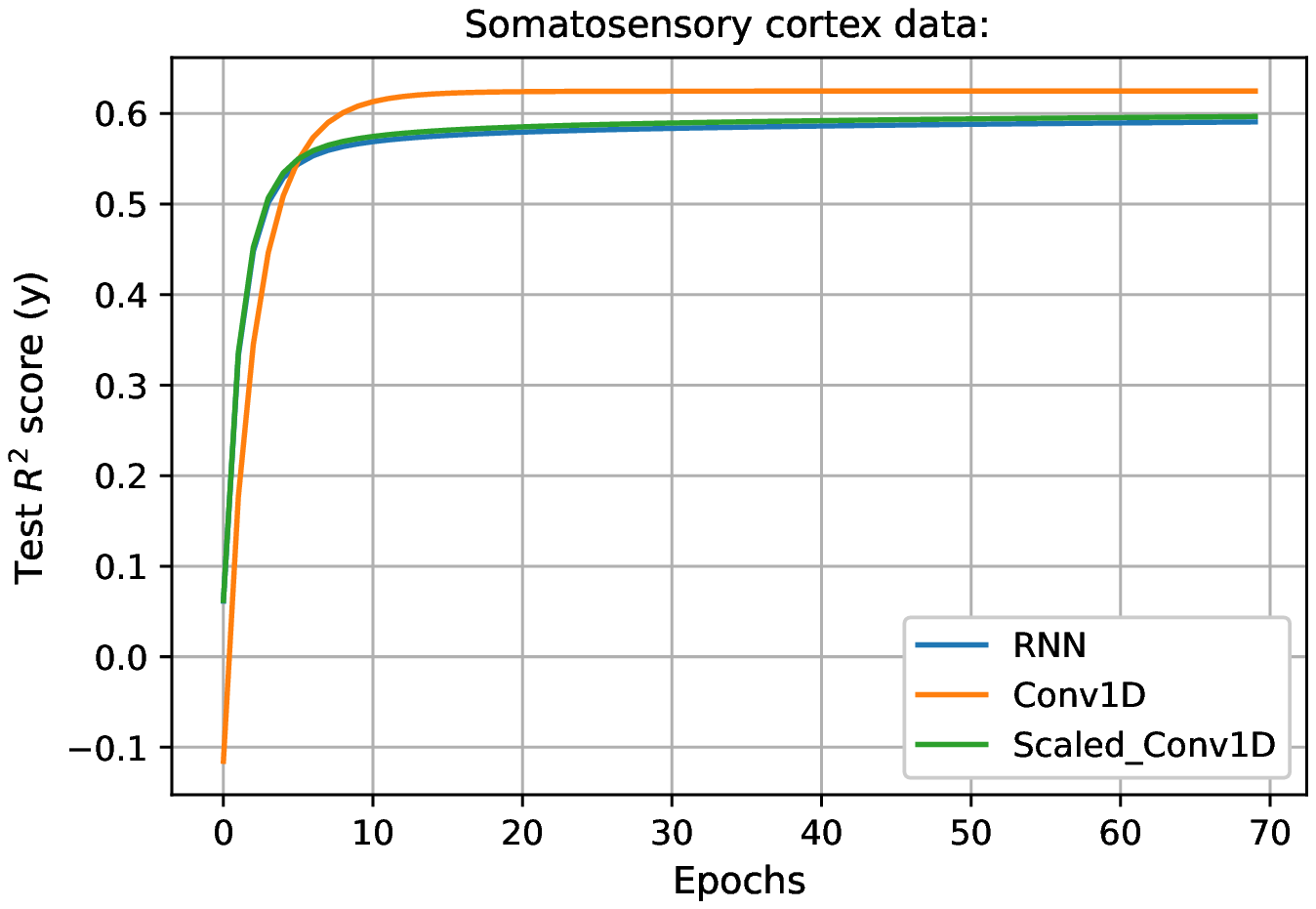}
\end{tabular}
\caption{$R^2$ score for the two dimensional reaching task described in section \ref{sec:real_data} . The data is recorded from the primary somatosensory cortex of macaques. (a) Limited data: The models are trained on 4.5 minutes of recorded data. (b) Entire data: the whole recording ($\approx 51$ mins) is used to compare the performances. For both cases we used mini-batch (batch size = 128) gradient descent with lr = $10^{-4}$. }
\label{fig:somato}
\end{figure}

Fig.~\ref{fig:somato} shows the $R^2$ scores for x and y directions of all three models for this task. Observe that, the dynamics of the linear RNN and scaled convolutional model are identical during training using either the entire recording or a part of it. For the case of limited data, as discussed earlier, we observe the implicit bias of the RNN and scaled convolutional model in the figures (a). This bias leads to better performance of these two models compared to the unscaled model. Using the total available data, the unscaled convolutional model performs better because of the increased bias error in the other two models (figures in (b)). Table \ref{tab:rsqr} shows the test $R^2$-score of the final trained models for all three cases.

\begin{table}
\begin{center}
 \begin{tabular}{c| c| c|c} 
   & RNN & Scaled Conv-1D & Conv-1D \\ 
 \hline\hline
 $R_x^2$ & 0.6462 & 0.6442  & 0.6565 \\ 
 \hline
 $R_y^2$ & 0.5911 & 0.5860  & 0.6027  \\
  \hline\hline
 $R_x^2$ \tiny{(limited data)} & 0.6043 & 0.6046  & 0.5856 \\ 
 \hline
 $R_y^2$ \tiny{(limited data)} & 0.4257 & 0.4234  & 0.3918  \\
\end{tabular}
\caption{$R^2$-score on test data for x and y directions in the two dimensional reaching task described in section \ref{sec:real_data}}
\label{tab:rsqr}
\end{center}
\end{table}

\section{Conclusion}

In this work, we focus on the special class of linear RNNs and observe a functional equivalence between linear RNNs and 1D convolutional models.
Using the kernel regime framework, we show that the training of a linear RNN is identical to the training of a certain scaled convolutional model. We further provide an analysis for an inductive bias in linear RNNs towards short-term memory. We show that this bias is driven by the variances of RNN parameters at random initialization. 
Our theory is validated by both synthetic and real data experiments. 

\bibliography{ref}
\bibliographystyle{alpha}

\newpage
\onecolumn
\appendix
\section{Background on Convergence of Vector Sequences and Random Variables}
\label{app:definitions}
In this section we review some the background on convergence of random variables, definitions of convergence of matrix sequences and some of their properties that we use throughout this paper.

\begin{definition}[Pseudo-Lipschitz continuity]
\label{def:pl_cont}
For a given $p\geq 1$, a function $f:\Real^{d}\rightarrow \Real^{m}$ is called pseudo-Lipschitz of order $p$, denoted by PL($p$), if
\begin{equation}
     \|f(\xbf_1)-f(\xbf_2)\|
    \leq C\|\xbf_1-\xbf_2\|\left(1+\|\xbf_1\|^{p-1}+\|\xbf_2\|^{p-1}\right)
\end{equation}
for some constant $C > 0$.  
\end{definition}
This is a generalization of the standard definition of Lipshitiz continuity. A PL(1) function is Lipschitz with constant $3C$.

\newcommand{\PLT}{\rm PL(2)}
\begin{definition}[Empirical convergence of a sequence]
\label{def:plp_convergence} Consider a sequence of vectors $\xbf(N) = \{\xbf_n{(N)}\}_{n=1}^N$ with $\xbf_n(N)\in\Real^{d}$. So, each $\xbf(N)$ is a block vector
with a total of $Nd$ components.  
For a finite $p\geq 1$, we say that the vector sequence $\xbf(N)$ converges empirically with $p$-th order moments if there exists a random variable $X \in \Real^d$ such that
\begin{enumerate}[label=(\roman*)]
\item $\Exp\|X\|_p^p < \infty$; and
\item for any $f : \Real^d \arr \Real$ that is pseudo-Lipschitz continuous of order $p$,
\beq \label{eq:PLp-empirical}
    \lim_{N \arr \infty} \frac{1}{N} \sum_{n=1}^N f(\xbf_n(N)) = \Exp\left[ f(X) \right].
\eeq
\end{enumerate}
\end{definition}

In this case, with some abuse of notation, we will write
\beq \label{eq:plLim}
    \lim_{n \arr \infty}  \xbf_n  \stackrel{PL(p)}{=} X,
\eeq
where we have omitted the dependence on $N$ in $\xbf_n(N)$.
We note that the sequence $\{\xbf{(N)}\}$ can be random or deterministic. If it is random, we will require that
for every pseudo-Lipschitz function $f(\cdot)$,
the limit \eqref{eq:PLp-empirical} holds almost surely.
In particular, if $\xbf_n \sim X$ are i.i.d.\ and $\Exp\|X\|^p_p < \infty$, then $\xbf$ empirically converges to $X$ with $p^{\rm th}$ order moments. 

Weak convergence (or convergence in distribution) of random variables is equivalent to
\begin{equation}
    \lim_{n\arr \infty}\Exp f(X_n) = \Exp f(X), \quad \text{for all bounded functions }f.
\end{equation}
It is shown in \cite{bayati2011dynamics} that PL$(p)$ convergence is equivalent to weak convergence plus convergence in $p^{\rm}$ moment.

\begin{definition}[Wasserstein-2 distance] Let $\nu$ and $\mu$ be two distributions on some Euclidean space $\Xmc$. The Wasserstein-2 distance between $\nu$ and $\mu$ is defined as
\begin{equation}
    W_2(\nu, \mu) = \left(\inf_{\gamma \in \Gamma} \Exp \norm{X - X'}_2^2\right)^{\frac{1}{2}},
\end{equation}
where $\Gamma$ is the set of all distributions with marginals consistent with $\nu$ and $\mu$. 
\end{definition}

A sequence $x_n$ converges PL(2) to $X$ if and only if the empirical measure $\hat{\Prob}_N = \frac{1}{N}\sum_{n=1}^N \delta(x-x_n)$ (where $\delta(\cdot)$ is the Dirac measure,) converges in Wasserstein-2 distance to distribution of $X$ \cite{villani2008optimal}, i.e.
\begin{equation}
    x_n \PLeq  X \quad \iff \quad \lim_{n\arr \infty} W_2(\hat{\Prob}_N, \Prob_X) = 0.\label{eq:PL2_W2_equivalence}
\end{equation}

For two zero mean Gaussian measure $\nu =  \Norm(0, \Sigma_1), \mu= \Norm(0, \Sigma_2)$ the Wasserstein-2 distance is given by \cite{givens1984class}
\begin{equation}
    W^2_2(\nu, \mu) = 
    \tr(\Sigma_1 - 2(\Sigma_1^{1/2}\Sigma_2\Sigma_1^{1/2})^{1/2}+\Sigma_2).
\end{equation}
Therefore, for zero mean Gaussian measures, convergence in covariance, implies convergence in Wasserstein-2 distance, and hence if the empirical covariance of a zero mean Gaussian sequence $x_n$ converges to some covaraince matrix $\Sigma$, then using \eqref{eq:PL2_W2_equivalence} $x_n \PLeq X$ where $X\sim \Norm(0, \Sigma)$.

\section{Proofs}

\subsection{Proof of Proposition \ref{prop:lin-RNN_equivalence} } 
\label{sec:equiv_proof}

Suppose we are given a convolutional model \eqref{eq:yconv}
with impulse response coefficients $L_t$, $t=0,\ldots,T-1$.
It is well-known from linear systems theory \cite{kailath1980linear} that 
linear time-invariant systems are input-output equivalent if and only if
they have the same impulse response coefficients.  So, we simply need to find
matrices $(W,F,C)$ satisfying \eqref{eq:linrnn_impulse}.
First consider the single input single output (SISO) case where 
$n_x= n_y=1$.  Take any set of 
real non-zero scalars $\lambda_i$, $i=0,\ldots,T-1$, that are distinct and set
\begin{equation}  \label{eq:Wequiv}
    W = \mathrm{diag}(\lambda_0,\ldots,\lambda_{T-1}), 
    \quad 
    F = 1_{T},
\end{equation}
so there are $n=T$ hidden states.
Then, for any $t$,
\begin{equation}
    (CW^tF) = \sum_{k=0}^{T-1} C_{k}\lambda_k^t.
\end{equation}
Equivalently, the impulse response coefficients in \eqref{eq:linrnn_impulse} are given by,
\begin{equation}
    [L_0, \cdots, L_{T-1}] = CV,
\end{equation}
where $V$ is the Vandermode matrix
$V_{jt} = \lambda_j^t$.  Since the values $\lambda_j$ are distinct, $V$
is invertible and we can find a vector $C$ matching arbitrary impulse response
coefficients.  Thus, when $n_x=n_y=1$, we can find a linear RNN with at most
$n=T$ hidden states that match the first $T$ impulse response coefficients.
To extend to the case of arbitrary $n_x$ and $n_y$, we simply create $n_xn_y$
systems, one for each input-output component pair.  Since each system
will have $T$ hidden states, the total number of states would be 
$n=Tn_xn_y$.

\subsection{Proof of Theorem \ref{thm:lin_param}}\label{app:proof_lin_param}
\newcommand{\sumjz}{\sum_{j=0}^{t}}
\newcommand{\sumkz}{\sum_{k=0}^{t}}

Given $y_t = \sumjz \sqrt{\rho_j} \theta_j x_{t-j}$ and $\theta = (\theta_0, \dots, \theta_{T-1})$, we consider a perturbation in $\theta$, namely $\Delta_\theta$. Therefore, 
\begin{align}
    \yt_t = \sumjz \sqrt{\rho_j} {\Delta_\theta}_j x_{t-j}
\end{align}
and the NTK for this model is given by
\begin{align}
    K_{t,s}(x, x') = \sum_{\Delta_\theta \in T_\theta} \yt_t(\Delta_\theta)\yt_s'(\Delta_\theta)\tran.
\end{align}
where $T_\theta$ is the standard basis for the parameter space. The following lemma shows this sum can be calculated as an expectation over a Gaussian random variable.
\begin{lemma}\label{lem:sum_to_Gaussian_exp}
Let $V$ be a finite dimensional Hilbert space and $W=\Real^m$ with the standard inner product and let $T, T':V\rightarrow W$ be linear transformations. Let $\{v_i\}_{i=1}^n$ be an ordered orthonormal basis for V. Then we have 
\begin{equation}
    \sum_{i=1}^n T(v_i)(T'(v_i))\tran = \Exp_{\alpha \sim \Norm(0, I_n)}
    \left[T\big(\sum_{i=1}^n\alpha_i v_i\big)\bigg(T'\big(\sum_{i=1}^n\alpha_i v_i\big)\bigg)\tran\right].
\end{equation}
\begin{proof}
\begin{align}
    \Exp_{\alpha \sim \Norm(0, I_n)}
    \left[T\big(\sum_{i=1}^n\alpha_i v_i\big)\bigg(T'\big(\sum_{j=1}^n\alpha_j v_j\big)\bigg)\tran\right] 
    &= \Exp_{\alpha \sim \Norm(0, I_n)}
    \left[\sum_{i, j=1}^n \alpha_i \alpha_j T\big( v_i\big)T'\big(v_j\big)\tran\right] \nonumber \\
    &=\sum_{i=1}^n T(v_i)(T'(v_i))\tran.
\end{align}
\end{proof}
\end{lemma}

Since $\yt_t(\Delta_\theta)$ is a linear operator, by applying Lemma \ref{lem:sum_to_Gaussian_exp} we have,
\begin{align}
    K_{t,s}(x, x') &= \Exp_{\Delta_\theta \sim \Norm(0,1), \text{i.i.d.}} \ [\yt_t(\Delta_\theta)\yt_s'(\Delta_\theta)\tran] \nonumber \\
    &=  \Exp_{\Delta_\theta \sim \Norm(0,1), \text{i.i.d.}} \ \left[(\sumjz \sqrt{\rho_j} {\Delta_\theta}_j x_{t-j}) ( \sumkz \sqrt{\rho_k} {\Delta_\theta}_k x'_{s-k})\tran\right] \nonumber \\
    &= \Exp_{\Delta_\theta \sim \Norm(0,1), \text{i.i.d.}} \ \left[(\sumjz \rho_j \ {\Delta_\theta}_j x_{t-j} {x'_{s-j}}\tran {\Delta_\theta}_j \tran )\right]
    \nonumber 
\end{align}
Therefore,
\begin{align*}
    \bigg(K_{t,s}(x, x')\bigg)_{m,m'} &= \Exp_{\Delta_\theta \sim \Norm(0,1), \text{i.i.d.}} \ \left[\sumjz \rho_j\ 
    \sum_{k,k'}{({\Delta_\theta}_j)}_{m,k} x_{t-j,k} {x'_{s-j,k'}} {({\Delta_\theta}_j)}_{k',m'}\right]\\
    &= (\sumjz \rho_j\ x_{t-j}\tran x'_{s-j}) \delta_{m,m'}
\end{align*}
Thus, 
$K_{t,s}(x, x') = (\sumjz \rho_j\ x_{t-j}\tran x'_{s-j} ) I_{n_y}$ and we can write the full kernel as 
\begin{align}
    K(x,x') = \Tc(x)\tran D(\rho)\Tc(x') \otimes I_{n_y},
\end{align}
where $\Tc(x)$ and $D(\rho)$ are defined in \eqref{eq:Txdef} and \eqref{eq:Dalpha} respectively. %
\subsection{Proof of Theorem \ref{thm:linrnn}} \label{app:linrnn_proof}
Part (a) is a special case of a more general lemma, Lemma \ref{lm:G-recursions} which we present in Appendix \ref{app:G-recursion}.
Let 
\begin{equation} \label{eq:linq}
    q_{t+1} = \frac{1}{\sqrt{n}} Wq_{t}, \quad q_0 = F,
\end{equation}
so that $q_t$ represents the impulse response from $x_t$ to $h_t$.
That is, 
\begin{equation} \label{eq:linqconv}
    h_t = \sum_{j=0}^t q_{t-j}x_j,
\end{equation}
which is the convolution of $q_t$ and $h_t$.
The system \eqref{eq:linq} is a special case of \eqref{eq:qmatrec1} with
$L=1$, no input $u_t$ and
\[
    A = W, \quad G(q) = q.
\]
Since there is only $L=1$ transform, we have dropped the dependence on the index
$\ell$.
Lemma \ref{lm:G-recursions}
hows that $(q_0,\ldots,q_t)$ converges $PL(2)$ to a Gaussian vector
$(Q_0,\ldots,Q_t)$ with zero mean. 
We claim that the $Q_i$'s are independent.
We prove this with induction.  Suppose $(Q_0,\ldots,Q_t)$ are independent.
We need to show $(Q_0,\ldots,Q_{t+1})$ are independent by 
using the SE equations \eqref{eq:gense}.
Specifically, from \eqref{eq:gense_z}, $Z_i = Q_i$ for all $i$.
Also, since each $Q_i$ is zero mean, $\mu_i = 0$ and $\Zt_i = Z_i=Q_i$.
Since the $\Zt_i$ are independent, the linear predictor coefficients
in \eqref{eq:gense_f} are zero:  $F_{ti}=0$.
Therefore, $\Rt_t=R_t \sim \Norm(0,\nu_W P_t)$ is independent of $(R_0,
\ldots,R_{t-1})$.  From \eqref{eq:gense_q}, $Q_{t+1}=R_t$.  So,
we have that $(Q_0,\ldots,Q_{t-1})$ is an independent Gaussian vector.
Finally, to compute the variance of the $Q_{t+1}$,   observe
\begin{align}
    \cov (Q_{t+1}) &\stackrel{(a)}{=} \cov(R_t)
        \stackrel{(b)}{=} \nu_W P_t  \nonumber \\
        & \stackrel{(c)}{=} \nu_W \cov(Z_t)
        \stackrel{(d)}{=} \nu_W \cov(Q_t),
\end{align}
where (a) follows from \eqref{eq:gense_q};
(b) follows from \eqref{eq:gense_rt} and the fact that $F_{ti}=0$
for all $i$; (c) follows from \eqref{eq:gense_p}; and (d) follows from the 
fact that $Z_t = Q_t$.
Also, since $q_0 = F$, it follows that $Q_0 \sim \Norm(0,\nu_F I)$.
We conclude that $\cov(Q_t) = \nu_F \nu_W^t I_{n_x}$. This proves part (a).

For part (b), we consider  perturbations $\Delta_W$, $\Delta_F$, and $\Delta_C$
of the parameters $W$, $F$, and $C$.  We have that,
\begin{equation} \label{eq: lin_rnn_der}
    \htil_t = \frac{1}{\sqrt{n}} W\htil_{t-1} + \frac{1}{\sqrt{n}} \Delta_W h_t + \Delta_F x_t, \qquad \yt_t = \frac{1}{\sqrt{n}}C \htil_t + \Delta_C h_t 
\end{equation}
Combining this equation with \eqref{eq:linrnn}, we see that 
the mapping from $x_t$ to $[h_t~\htil_t]$ is a linear time-invariant
system.  Let $q_t \in \Real^{n \times 2 n_x}$ 
be its impulse response.  The impulse response coefficients
satisfy the recursive equations,
\[
    q_{t+1} = \left[\frac{1}{\sqrt{n}} Wq_{t,1},  \frac{1}{\sqrt{n}}(W q_{t,2} + \Delta_W q_{t,1})\right],
    \quad q_0 = \left[ F,  \Delta_F \right].
\]
We can analyze these coefficients in the LSL using 
Lemma \ref{lm:G-recursions}.  Specifically, let $L=2$ and set
\[
    A_1 = W, \quad A_2 = \Delta_W.
\]
Also, let
\begin{subequations}  \label{eq:linntk_z}
\begin{align} 
     z_{t1} &= \overline{G}_1(q_t) := q_t, \\
     z_{t2} &= \overline{G}_2(q_t) := [0\  ~ q_{t1}].
\end{align}
\end{subequations}
Then, we have the updates,
\[
    q_{t+1} = \frac{1}{\sqrt{n}} Wz_{t1} + \frac{1}{\sqrt{n}} \Delta_W z_{t2}, \quad q_0 = \left[ F,  \Delta_F \right].
\]
It follows from Lemma~\ref{lm:G-recursions} that
$(q_0,\ldots,q_{T-1})$ converges $PL(2)$ to zero mean
Gaussian random variables $(Q_0,\ldots,Q_{T-1})$. Note that $Q_t = [Q_{t1}, Q_{t2}]$ where each $Q_{t1}$ and $Q_{t2}$ are random vectors $\in \Real^{1\times n_x}$.

Similar to the proof of the previous theorem, we use induction to show 
that $(Q_0,\ldots,Q_{t})$ are independent.  Suppose that the claim
is true for $t$.  Then, $Z_{i1}$ and $Z_{i2}$ are functions
of $Q_i$.  So, for $\ell=1,2$, $Z_{t\ell}$ is independent of
$Z_{i\ell}$ for $i<t$.   Thus, the prediction coefficients $F_{ti\ell}=0$
and, as before, $R_{t\ell} \sim \Norm(0,P_{t\ell})$ independent of 
$R_{i\ell}$, $i < t$.  Thus, $Q_{t+1} = R_{t1} + R_{t2}$ is independent of
$(Q_0,\ldots,Q_t)$.

We conclude by computing the $\cov(Q_t)$.  
We claim that, for all $t$, the variance of $Q_t$ is of the form,
\beq \label{eq:varQ_linntk}
    \cov(Q_t) = \left[ \begin{array}{cc}  \tau_{t1} I_{n_x} & 0 \\ 0 & \tau_{t2} I_{n_x} \end{array}\right]
\eeq
for scalar $\tau_{t1}, \tau_{t2}$.  Since $q_0 = \left[ F,  \Delta_F \right]$,
we have 
\[
    \tau_{t1} = \nu_F, \quad \tau_{t2} = 1.
\]
Now suppose that \eqref{eq:varQ_linntk} is true for some $t$.
From \eqref{eq:gense_z}, 
\[
    Z_{t1} = Q_t, \quad Z_{t1} = (0, Q_{t1}),
\]
from which we obtain that 
\beq \label{eq:varz_linntk}
    \cov (Z_{t1}) = \left[ \begin{array}{cc}  \tau_{t1}I_{n_x} & 0 \\ 0 & \tau_{t2} I_{n_x} \end{array}\right],
    \quad
    \cov(Z_{t2}) = \left[ \begin{array}{cc}  0 & 0 \\ 0 & \tau_{t1}I_{n_x} \end{array}\right].
\eeq
Therefore, we have
\begin{align}
    \cov(Q_{t+1}) &\stackrel{(a)}{=} \cov(R_{t,1})
    + \cov(R_{t,2}) \nonumber \\
        & \stackrel{(b)}{=} \nu_W P_{t1} + P_{t2}   \nonumber \\
        & \stackrel{(c)}{=} \nu_W \cov(Z_{t1}) 
         + \cov(Z_{t2})   \stackrel{(d)}{=} 
         \left[ \begin{array}{cc}  \nu_W \tau_{t1} I_{n_x} & 0 \\ 0 & (\tau_{t1}  + 
         \nu_W \tau_{t2})I_ {n_x} \end{array}\right],
\end{align}
where (a) follows from \eqref{eq:gense_q};
(b) follows from \eqref{eq:gense_rt} and the fact that $F_{ti\ell}=0$
for all $i$; (c) follows from \eqref{eq:gense_p}; and (d) follows from 
\eqref{eq:varz_linntk}.  It follows that
\[
    \tau_{t+1,1} = \nu_W \tau_{t1}, \quad
    \tau_{t+1,2} = \nu_W \tau_{t2} + \tau_{t1}.
\]
These recursions have the solution,
\begin{equation}\label{eq:rec_sol_tau}
    \tau_{t1} = \nu_W^t \nu_F, \quad
    \tau_{t2} = t\nu_F\nu_W^{t-1} + \nu_W^t.
\end{equation}

Since $ [h_t, \htil_t] = \sumjz q_{j} \begin{bmatrix}x_{t-j} \\ x_{t-j} \end{bmatrix} $ and we know each $q_t$ converges $PL(2)$ to random $Q_t$ with covariances calculated in \eqref{eq:varQ_linntk} and \eqref{eq:rec_sol_tau}, we have
\begin{align}
   [h_t, \htil_t] \PLeq [H_t,\Ht_t] = \sumjz Q_{j} \begin{bmatrix}x_{t-j} \\ x_{t-j} \end{bmatrix}
\end{align}
where $H_t, \Ht_t$ are scalar random variables. For each $t, s$, we can now calculate the auto-correlation function for $H$ as follows
\begin{align}\label{eq:EH}
    \Exp [H_t H_s] & = \Exp[\sumjz \sumkz x_{t-j}\tran Q_{j,1}\tran Q_{k,1} x_{s-k}]\nonumber \\
    & = \sumjz x_{t-j}\tran \Exp[Q_{j,1}\tran Q_{j,1}] x_{s-j}\nonumber \\
    &= \sumjz \nu_W^j\nu_F\  x_{t-j}\tran x_{s-j}.
\end{align}
Similarly for $\Ht$ we have 
\begin{align}\label{eq:EHt}
    \Exp [\Ht_t \Ht_s] = \sumjz\ (j\nu_F \nu_W^{j-1} + \nu_W^j)\ x_{t-j}\tran x_{s-j}.
\end{align}
Thus, the impulse response of the system $L_j = C q_{j,1}$ converge empirically to $\Norm(0, \Lambda)$ where,
\begin{align}
    \Lambda = \nu_C \nlim \frac{1}{n}  q_{j,1}\tran q_{j,1} = \nu_C \Exp [Q_{j,1}\tran Q_{j,1}] =  \nu_C \nu_F \nu_W^j I_{n_x}.
\end{align}
This proves part (a).

Note that $\Exp [\Ht_t \Ht'_s]$ and $\Exp[H_t H'_s]$ can be calculated similarly by substituting $x_{s-j}$ with $x'_{s-j}$ in \eqref{eq:EH} and \eqref{eq:EHt}.

Next, we calculate the NTK in this case
\begin{align}
     K_{t,s}(x, x') &= \sum_{\Delta_\Theta \in T_\Theta} \yt_t(\Delta_\Theta)\yt_s'(\Delta_\Theta)\tran \nonumber \\
     &\stackrel{(a)}{=}\Exp_{\Delta_\Theta \sim \Norm(0,1), \text{i.i.d.}} \ [\yt_t(\Delta_\Theta)\yt_s'(\Delta_\Theta)\tran]
\end{align}
where (a) follows from Lemma \ref{lem:sum_to_Gaussian_exp}. 
Combining with \eqref{eq: lin_rnn_der} we have
\begin{align}
   K_{t,s}(x, x') &= \Exp_{C, \Delta_C \sim \Norm(0,1), \text{i.i.d.}} \ \left[(C\htil_t + \Delta_C h_t)(C \htil'_s + \Delta_C h_s')\tran \right]\nonumber \\
   &= \left(\nu_C \Exp[\Ht_t \Ht'_s] + \Exp[H_t H'_s]\right) I_{n_y}
\end{align}
Therefore, 
\begin{align}
    K(x,x') = \Tc(x)\tran D(\rho)\Tc(x') \otimes I_{n_y},
\end{align}
where $\Tc(x)$ and $D(\rho)$ are given in \eqref{eq:Txdef}
and \eqref{eq:Dalpha} and 
\beq
    \rho_i =\nu_C (i\nu_F\nu_W^{i-1} + \nu_W^i) + \nu_W^i \nu_F.
\eeq
This proves part (b).

\subsection{Proof of Theorem \ref{thm:implicit_bias}} \label{app:implicit_bias_proof}

\paragraph*{Bounding the Initial Impulse Response}
from Theorem~\ref{thm:linrnn},
each coefficient of $L_{{\rm RNN},j}^0$
has mean zero and variance $\nu_C\nu_F\nu_W^j$.  There are $n_xn_y$
such components. This proves \eqref{eq:Lrnn_norm_init}.

\paragraph{Convolutional Equivalent
Linear Model}
The key for the remainder of the proof is to use
Theorems~\ref{thm:lin_param} and \ref{thm:linrnn} to
construct a scaled convolutional
model that has the same NTK and intial conditions 
as the RNN.  Then, we analyze the convolutional
model to obtain the desired bound.
To this end, 
let $\mbb{\rho} = [\rho_0,\ldots,\rho_{T-1}]$ be
the scaling factors given in Theorem~\ref{thm:linrnn}.
For each initial condition $\theta^0_{\rm RNN}=(W^0,F^0,C^0)$
of the RNN, suppose that we initialize the
scaled convolutional model with 
\[
    \theta_{{\rm conv},j}^0 = 
    \frac{1}{\sqrt{\rho_j} n^{(j+1)/2}} C^0(W^0)^j F^0.
\]
The initial impulse response of the 
scaled convolutional model will then be
\begin{equation} \label{eq:lconv_init}
    L_{{\rm conv},j}^0 = \sqrt{\rho_j}
    \theta_{{\rm conv},j} = 
    \frac{1}{n^{(j+1)/2}} C^0(W^0)^j F^0 = 
    L_{{\rm RNN},j}^0.
\end{equation}
Hence, the scaled convolutional model and the RNN
have the same initial impulse response coefficients.
We then train the scaled convolutional model on the training
data using gradient descent with the same learning rate
$\eta$ used in the training of the RNN.  
Let $L_{{\rm conv},j}^\ell$
denote the impulse response of the scaled convolutional
model after $\ell$ steps of gradient descent.

\paragraph*{Gradient Descent Analysis of
the Convolutional Model} 
Next, we look at how
the impulse response of the 
scaled convolutional model evolves over the gradient
descent steps.  It is convenient to do this analysis
using some matrix notation.
For each parameter,
$\theta=[\theta_0,\ldots,\theta_{T-1}]$, the
convolutional filter parameters are $L_j = \sqrt{\rho_j} \theta_j$.
Thus, we can write
\[
    \mathbf{L} = \Db^{1/2}\theta,
\]
where $\Db$ is a block diagonal operator with values $\rho_j$.
Also, let $\wh{\yb} = [\wh{\yb}_1,\ldots,\wh{\yb}_N]$ be the set of predictions
on the $N$ training samples.  Since the convolutional model is linear, we
can write $\hat{\yb} = \Ab \mathbf{L}$ for some linear operator $\Ab$.  The operator
$\Ab$ would be a block Toeplitz with the input data $\xb=[\xb_1,\ldots,\xb_N]$.
Also, if we let $\yb=[\yb_1,\ldots,\yb_N]$ be
the $N$ training samples, the least squares 
cost is
\[
    \|\yb - \Ab \Db^{1/2}\thb \|^2_F.
\]
Minimizing this loss function 
will result in GD steps,
\[
    \theta^{\ell+1} = \theta^\ell + \eta \Db^{1/2}
    \Ab\tran (\yb - \Ab \Db^{1/2} \theta^\ell).
\]
Now let $\ub^\ell = \Db^{-1/2}(\thb^\ell - \thb^0)$
and $\bb := \yb - \Ab\Db^{1/2}\thb^0$.
Then,
\begin{equation}
    \ub^{\ell+1} = \ub^\ell + \eta \Ab\tran 
    (\bb - \Ab \Db)\ub^\ell
    = (\I - \eta \Ab\tran\Ab\Db)\ub^\ell
    + \eta\Ab\tran \bb \label{eq:uupdate1}
\end{equation}
For $0 < \nu_W < 1$, we have that $\rho_j$ satisfies the
bound \eqref{eq:rhodecay}.
Since $\Db$ is a block diagonal matrix
with entries $\rho_j$, 
$\|\Db\| \leq \rho_{\rm max}$.
Now select 
\begin{equation}
    B_1 := \frac{1}{\rho_{\rm max}\|\Ab\|^2},
    \quad
    B_2 := \|\Ab\tran\bb\|.
\end{equation}
If we take $\eta < B_1$ then 
\[
    \eta\Db^{1/2}\Ab\tran \Ab\Db^{1/2} \leq 
    \eta \rho_{\rm max}\|\Ab\|^2 \leq \I 
    \Rightarrow
    \| \Ib - \eta \Ab\tran\Ab\Db \| \leq 1.
\]
Hence, \eqref{eq:uupdate1} shows that 
\begin{equation} \label{eq:ubnd}
    \|\ub^{\ell+1}\|_F \leq  \|\ub^\ell\|_F + \eta B_2
    \Rightarrow \|\ub^\ell\|_F \leq \eta \ell B_2,
\end{equation}
where we have used the fact that $\ub^0 = \mathbf{0}$.
Now, since $\ub^\ell = \Db^{-1/2}(\thb^\ell - \thb^0)$,
the $j$-th component of $\thb^\ell$ is
\[
    \thb^\ell_j = \thb^0_j + \sqrt{\rho_j}\ub^\ell_j.
\]
Hence,
\begin{align*}
    L_{{\rm conv},j}^\ell &= \sqrt{\rho_j}\theta_{j}^\ell
    = L_{{\rm conv},j}^0 + \rho_j \ub^\ell_j.
\end{align*}
Applying \eqref{eq:ubnd} we obtain the bound on the convolutional model
\begin{equation} \label{eq:Lconv_bnd}
    \|L_{{\rm conv},j}^\ell-L_{{\rm conv},j}^0\|_F
    \leq \rho_j \eta \ell B_2.
\end{equation}

\paragraph*{Bounding the RNN Impulse Response}
From Theorems~\ref{thm:lin_param} and \ref{thm:linrnn},
the scaled convolutional model and linear RNN
have the same NTK.  Due to \eqref{eq:lconv_init},
they have the same input-output mapping at the initial conditions.
Since the scaled convolutional model is linear in its
parameters it follows that it is linear NTK model for
the RNN.  Therefore, using the NTK results such as
Proposition~\ref{prop:ntk_conv}, we have that
for all input sequences $\xb$ and GD time steps $\ell$,
\begin{equation} \label{eq:conv_rnn_equiv}
    \lim_{n \rightarrow \infty} 
    \left \| f_{\rm RNN}(\xb,\theta^\ell_{\rm RNN}) -
    f_{\rm conv}(\xb,\theta^\ell_{\rm conv}) \right\| = 0,
\end{equation}
where the convergence is in probability.
Thus, if we fix an input $\xbf$ and iteration $\ell$ and define
\[
    \yb_{\rm RNN} = f_{\rm RNN}(\xbf,\theta_{\rm RNN}^\ell), \quad
    \yb_{\rm conv} = f_{\rm conv}(\xbf,\theta_{\rm conv}^\ell),
\]
the limit \eqref{eq:conv_rnn_equiv} can be re-written as
\begin{equation} \label{eq:y_rnn_equiv}
    \lim_{n \rightarrow \infty} 
    \left \| y_{{\rm RNN}, j}-  y_{{\rm conv}, j} \right\| = 0,
\end{equation}
for all $j$.  Again, the convergence is in probability.  Now consider the case 
where the input sequence $\xbf=(x_0,\ldots,x_{T-1})$ is a sequence with $x_j = 0$
for all $j > 0$   That is, it is only non-zero at the initial time step.
Then for all time steps $y_{{\rm RNN}, j} = L_{{\rm RNN}, j}^\ell x_0$ and 
$y_{{\rm conv}, j} = L_{{\rm conv}, j}^\ell x_0$.  Since this is true for all $x_0$,
\eqref{eq:y_rnn_equiv} shows that for all time steps $j=0,\ldots,T-1$,
\begin{equation} \label{eq:conv_rnn_imp}
    \lim_{n \rightarrow \infty} 
    \left \|  L_{{\rm RNN},j}(x,\theta^\ell_{\rm RNN}) -
    L_{{\rm conv},j}(x,\theta^\ell_{\rm conv}) \right\|_F = 0
\end{equation}
where the convergence is in probability.
Combining \eqref{eq:conv_rnn_imp} with \eqref{eq:Lconv_bnd} proves \eqref{eq:Lrnn_norm_iter}. %
\newpage
\section{Recursions with Random Gaussians}\label{app:G-recursion}

We consider a recursion of the form,
\beq \label{eq:qmatrec1}
    q_{t+1} = \sum_{\ell=1}^L \frac{1}{\sqrt{n}} A_\ell 
        \overline{G}_\ell(q_t,u_t),
\eeq
where $q_t \in \Real^{n \times d_q}$, $u_t \in \Real^{n \times d_u}$, and
$\overline{G}_\ell(q_t,u_t)$ acts row-wise, meaning
\beq \label{eq:Grow}
    \overline{G}_\ell(q_t,u_t)_{i,:} = 
    G_\ell(q_{t,i,:},u_{t,i,:}),
\eeq
for some Lipschitz functions $G_{\ell}:\Real^{d_q}\times \Real^{d_u}\times \rightarrow\Real^{d_q}$.  That is, the outptut of 
row $i$ of $\overline{G}_\ell(\cdot)$ depends only the $i$-th
rows of its inputs.  We will analyze this system for a fixed horizon,
$t=0,\ldots,T-1$.  Assume that
\beq \label{eq:uconv1}
    (q_0,u_0,\ldots,u_{T-1}) \stackrel{PL(2)}{\rightarrow} (Q_0,U_0,\ldots,U_{T-1}),
\eeq
to random variables $(Q_0,U_0,\ldots,U_{T-1})$ where $Q_0$ is independent
of $(U_0,\ldots,U_{T-1})$, and $Q_0  \sim \mathcal{N}(0,P_0)$
for some covariance matrix $P_0 \in \Real^{d_q \times d_q}$.  
Assume the matrices $A_{\ell} \in \Real^{n \times n}$
are independent with i.i.d.\ components, $(A_{\ell})_{i,j} \sim \mathcal{N}(0,\nu_\ell)$.

\begin{lemma} \label{lm:G-recursions}
 Under the above assumptions,
 \begin{equation}
     (q_0, q_1, \dots, q_{T-1}, u_0, u_1, \dots ,u_{T-1}) \stackrel{PL(2)}{\rightarrow} (Q_0, Q_1, \dots, Q_{T-1}, U_0, \dots U_{T-1})
 \end{equation}
 where each $Q_i \in \Real^{d_q}$ and $(Q_0, Q_1, \dots, Q_{T-1})$ are zero mean Gaussian processes independent of $(U_0, \dots U_{T-1})$, generated recursively through SE equations given by
\begin{subequations} \label{eq:gense}
\begin{align}
    D_\ell &= {\mathcal{N}}(0,\nu_\ell) \\
    Z_{t \ell} &= G_\ell(Q_t, U_t)    \label{eq:gense_z}\\
    \mu_{t \ell} &= \Exp(Z_{t \ell}), \quad \tilde{Z}_{t \ell} = Z_{t \ell} - \mu_{t \ell} \\
    F_{t,:,\ell} &= \min_{F_1,\ldots,F_{t-1}}
    \Exp \left\|\tilde{Z}_{t \ell} - \sum_{j=1}^{t} \tilde{Z}_{t-j, \ell}F_j \right\|^2 
    \label{eq:gense_f}\\
    P_{t\ell} &= \Exp (\tilde{Z}_{t \ell} - \sum_{j=1}^{t} \tilde{Z}_{t-j, \ell}F_{t j \ell })\tran
        (\tilde{Z}_{t \ell} - \sum_{j=1}^{t} \tilde{Z}_{t-j,\ell}F_{t j \ell}) 
        \label{eq:gense_p} \\
    \tilde{R}_{t\ell} &= \sum_{j=1}^{t} \tilde{R}_{t-j,\ell}F_{t j \ell} 
    + {\mathcal{N}}(0, \nu_W P_{t\ell}), \label{eq:gense_rt} \\
    R_{t\ell} &= \tilde{R}_{t\ell} + D_\ell \mu_{t \ell} \label{eq:gense_r} \\
    \quad Q_{t+1} &= \sum_{\ell=1}^L R_{t\ell}  \label{eq:gense_q}
\end{align}
\end{subequations}
\end{lemma}

\paragraph{Proof:}

We prove this by induction. Let $\Mmc_t$ be the hypothesis that this result is true up to iteration $t$. We show that $\Mmc_0$ is true and that $\Mmc_t$ implies $\Mmc_{t+1}$. 

\paragraph{Base case ($\mc M_0$):} 
Define $z_{0\ell} = \overline{G}_\ell(q_0, u_0)$. 
We have that rows of $z_{0 \ell}$ converge $PL(2)$ to $Z_{0 \ell} = \Gbar_\ell(Q_0, U_0)$.

Now, let $\mu_{0\ell} = \Exp(Z_{0 \ell})$ and define the following:
\begin{align}
    d_\ell &= \frac{1}{\sqrt{n}}A_\ell \ones,  \quad
    \zt_{0 \ell} = z_{0 \ell } - \ones \mu_{0 \ell } \\
    \rt_{1\ell} &= \frac{1}{\sqrt{n}}A_\ell \zt_{0 \ell }, \quad
    r_{1\ell} = \rt_{1\ell} + d_\ell \mu_{0\ell}, \quad
    q_{1} = \sum_{\ell=1}^L r_{1\ell}.
\end{align}
We know that 
\begin{align}
    d_\ell \PLeq D_\ell \sim \Norm(0, \nu_\ell), \qquad \zt_{0\ell} \PLeq \Zt_{0\ell}=Z_{0\ell}-\mu_{0\ell}. 
\end{align}
Note that $\Zt_{0\ell}$ are zero mean. 
Now since $A_\ell$ are i.i.d\ Gaussian matrices, rows of $\rt_{1 \ell}$ converge PL(2) to random variable 
\beq
\Rt_{1\ell} \sim \Norm(0, P_{1\ell}) \qquad \text{where, }\quad P_{1\ell} = \nlim \frac{1}{n} {\zt_{0 \ell}} \tran \zt_{0 \ell} \stackrel{a.s.}{=} \Exp (\Zt_{0 \ell}\tran \Zt_{0 \ell})
\eeq
Furthermore, one can show that $\Exp(\wt R_{1\ell_1}\wt R_{1\ell_2})=\Exp (\Zt_{0 \ell_1}\tran \Zt_{0 \ell_2})=0$ $\wt R_{1\ell_1}$ and $\wt R_{1\ell_2}$ are independent
Therefore, 
\begin{align}
    q_{1} \PLeq Q_{1} = \suml \left[\Rt_{1\ell} + D_\ell \mu_{0\ell}\right]
\end{align}
This proves $\Mmc_0$ holds true. 

\paragraph{Induction recursion:}
We next assume that the SE system is true up to iteration $t$.
We write the recursions as
\begin{subequations} \label{eq:genvec}
\begin{align}
    d_\ell &= \frac{1}{\sqrt{n}}A_\ell \ones \in \Real^n \\
    z_{t\ell} &= \overline{G}_\ell(q_t, u_t),   \quad
    \zt_{t \ell} = z_{t\ell} - \ones \mu_{t\ell } \\
    \rt_{t+1,\ell} &= \frac{1}{\sqrt{n}}A_\ell \zt_{t\ell}, \quad
    r_{t+1,\ell} = \rt_{t+1,\ell} + d_\ell \mu_{t\ell}, \quad
    q_{t+1} = \sum_{\ell=1}^L r_{t+1,\ell}.
\end{align}
\end{subequations}
The main issue in dealing with a recursion of the form Equation \eqref{eq:genvec} is that for $t\geq 1$, matrices $\{A_\ell\}_{\ell=1}^L$ and $\{\rt_{t\ell}\}_{\ell=1}^L$ are no longer independent. The key idea is to use a  conditioning technique (Bolthausen conditioning) as in \cite{bayati2011dynamics} to deal with this dependence. Instead of conditioning $\rt_{t\ell}$ on $A_{\ell}$, we condition $A_\ell$ on the event
\beq\label{eq:conditioning_event}
\mcE_{t,\ell} = \{\rt_{t'+1, \ell} = \frac{1}{\sqrt{n}} A_\ell \zt_{t'\ell}, t' = 0, \dots, t-1\}.
\eeq
Note that this event is a set of linear constraints, and i.i.d.\ Gaussian random variables conditioned on linear constraints have Gaussian densities that we can track.

Let $\Hmct_{t\ell}$ be the linear operator
\beq\label{eq:Hmct_definition}
\Hmct_{t\ell}:A_\ell \mapsto (\rt_{1\ell}, \dots, \rt_{t\ell}).
\eeq
With these definitions, we have
\beq
A_\ell|_{\varepsilon_{t,\ell}} \eqd \Hmct_{t\ell}^\dagger(\rt_{1\ell}, \dots, \rt_{t\ell}) + \Hmct_{t\ell}^\perp(\At_\ell),
\eeq
where $\Hmct_{t,\ell}^\dagger$ is the Moore-Penrose pseudo-inverse operator of $\Hmct_{t,\ell}$, $\Hmct_{t,\ell}^\perp$ is the orthogonal projection operator onto the subspace orthogonal to the kernel of $\Hmct_t$, and $\At_\ell$ is an independent copy of $A_\ell$. Therefore, we can write $\rt_{t+1,\ell}$ as sum of two terms
\begin{align}
\rt_{t+1,\ell} = \rt_{t+1,\ell}^\dete + \rt_{t+1,\ell}^\ran,
\end{align}
where $\rt_{t+1,\ell}^\dete$ is what we call the deterministic part:
\begin{align}\label{eq:deterministic_part}
\rt_{t+1,\ell}^\dete = \frac{1}{\sqrt{n}}\Hmct_{t\ell}^\dagger(\rt_1, \dots, \rt_{t})\ \zt_{t\ell}
\end{align}
and $\rt_{t+1,\ell}^\ran$ is the random part:
\begin{align}\label{eq:random_part}
\rt_{t+1,\ell}^\ran = \frac{1}{\sqrt{n}}\Hmct_{t\ell}^\perp(\At_\ell)\ \zt_{t\ell}.
\end{align}
It is helpful to write the linear operators defined in this section in matrix form for derivations that follow.
\begin{align}
    \Hmct_{t\ell}(A_\ell) &= \frac{1}{\sqrt{n}}[\begin{matrix}A_\ell \end{matrix}]\left[\begin{matrix}\zt_{0\ell}&\dots & \zt_{t-1,\ell}\end{matrix}\right].
\end{align}
\paragraph{Deterministic part:}
We first characterizes the limiting behavior of $\rt_{t+1,\ell}^\dete$.

It is easy to show that if the functions $\overline{G}_\ell$ are non-constant, then the operator $\Hmct_{t\ell} \Hmct_{t\ell}\tran$ where $\Hmct_{t\ell}\tran$ is the adjoint of $\Hmct_{t\ell}$, is full-rank almost surely for any finite $t$. Thus, we have
\begin{equation}
    \Hmct_{t\ell}^\dagger = \Hmct_{t\ell}\tran(\Hmct_{t\ell} \Hmct_{t\ell}\tran)\inv
\end{equation}
Form equation \eqref{eq:Hmct_definition} we have
\begin{equation}\label{eq:Hmct_adjoint}
    \Hmct_{t\ell}\tran(\rt_{1\ell}, \dots, \rt_{t\ell}) = \frac{1}{\sqrt{n}}\sum_{t'=1}^t \rt_{t'\ell}(\zt_{t'-1,\ell})\tran. 
\end{equation}
Combining \eqref{eq:Hmct_adjoint} and \eqref{eq:Hmct_definition} we get
\begin{align}
     \left(\Hmct_{t\ell} (\Hmct_{t\ell}\tran)(\rt_{1\ell}, \dots, \rt_{t\ell})\right)_{s}
    &= \frac{1}{n} \sum_{t'=1}^t \rt_{t'\ell}\ (\zt_{t'-1,\ell})\tran\zt_{s-1,\ell}
\end{align}
Now, under the induction hypothesis, using the definition of PL(2) convergence we have
\begin{align}
  R_{\Zt \ell}(t',s) := \lim_{n\rightarrow \infty} \frac{1}{n}(\zt_{t'-1,\ell})\tran \zt_{s-1,\ell}   &\stackrel{a.s.}{=} \Exp\left((\Zt_{t'-1,\ell})\tran \Zt_{s-1,\ell}\right)
\end{align}
Therefore we have, 
\begin{equation}
    \Hmct_{t\ell} \Hmct_{t\ell}\tran(\rt_{1 \ell}, \dots, \rt_{t \ell}) = [\begin{matrix}\rt_{1 \ell} & \dots & \rt_{t \ell}
    \end{matrix}] \underbrace{\left[\begin{matrix} R_{\Zt \ell}(0,0) & R_{\Zt \ell}(0,1) & \dots & R_{\Zt \ell}(0, t-1)\\
     R_{\Zt \ell}(1,0) & R_{\Zt \ell}(1,1) & \dots & R_{\Zt \ell}(1, t-1)\\
     \vdots & \vdots & \ddots &\vdots\\
      R_{\Zt \ell}(t-1, 0) & R_{\Zt \ell}(t-1,1) & \dots & R_{\Zt \ell}(t-1, t-1)
      \end{matrix}\right]}_{\Rmc_{\Zt \ell}}
\end{equation}
Let $\Rmc\inv_{\Zt \ell}$ denote the inverse of $\Rmc_{\Zt \ell}$ and index its blocks similarly to $\Rmc_{\Zt \ell}$.
Then, the pseudo-inverse is
\begin{equation}
    \Hmct_{t\ell}^\dagger(\rt_{1 \ell}, \dots, \rt_{t \ell}) =  \frac{1}{\sqrt{n}}\sum_{t'=1}^t \sumtt \rt_{t'' \ell}\Rmc_{\Zt \ell}\inv(t''-1, t'-1)(\zt_{t'-1,\ell})\tran +o(\frac{1}{n}).
\end{equation}
Define 
$
    \Pt_{t\ell} := \Zt_{t\ell} - \sum_{j = 1}^{t} \Zt_{t-j, \ell} F_{t j \ell},
$
, where $F_{t,:,\ell}$ are defined in \eqref{eq:gense_f}. 
Using equation \eqref{eq:deterministic_part} we get:
\begin{subequations}
\begin{align}
    \rt_{t+1,\ell}^\dete 
    &= \frac{1}{n}\sumtt  \rt_{t'' \ell} \sumt\Rmc_{\Zt \ell}\inv(t''-1, t'-1)(\zt_{t'-1,\ell})\tran  \zt_{t, \ell}+o(\frac{1}{n})\\
    &\stackrel{a.s.}{=}\sumtt \rt_{t'' \ell} \sumt \Rmc_{\Zt \ell}\inv(t''-1, t'-1)\ \Exp \left((\Zt_{t'-1,\ell})\tran \Zt_{t,\ell}\right)+o(\frac{1}{n})\\
    &= \sumtt \rt_{t'' \ell} \sumt \Rmc_{\Zt \ell}\inv(t''-1, t'-1)\ \Exp\left((\Zt_{t'-1,\ell})\tran (\Pt_{t\ell}+\sum_{j=1}^{t} \Zt_{t-j,\ell}F_{t,j,\ell}) \right) +o(\frac{1}{n})\\
    &= \sumtt \rt_{t'' \ell} \sumj  \underbrace{ \sumt  \Rmc_{\Zt \ell}\inv(t''-1, t'-1)\  \Rmc_{\Zt \ell} (t'-1, t-j)}_{I \delta(t''= t-j+1)} F_{t,j,\ell}  +o(\frac{1}{n})\\
    &= \sumj \rt_{t-j+1, \ell} F_{t,j,\ell}+o(\frac{1}{n}),
    \label{eq:q_det_coefs}
\end{align}
\end{subequations}
where (a) follows from the fact that $ \Exp (\Zt_{t' \ell}\tran \Pt_{t\ell}) = 0$ for $t' = 0, \dots , t-1$.
Now by induction hypothesis we know that $\rt_{t-j+1,\ell} \PLeq \Rt_{t-j+1,\ell}$, therefore,
\begin{align}
    \rt_{t+1,\ell}^\dete \PLeq \Rt_{t+1,\ell}^\dete = \sumj \Rt_{t-j+1,\ell} F_{t,j,\ell}
\end{align}

\paragraph{Random part}
We next consider the random part:

\begin{align}
    \rt_{t+1,\ell}^\ran &= \frac{1}{\sqrt{n}}\Hmct_{t\ell}^\perp(\At_\ell) \zt_{t\ell}\\
    &= \frac{1}{\sqrt{n}} (\At_\ell \zt_{t\ell} - \Hmct_{t\ell}^\dagger \Hmct_{t\ell}(\At_\ell)\zt_{t\ell}).
\end{align}
We know that,
\begin{align}
    \Hmct_t^\dagger \Hmct_t(\At_\ell) = \frac{1}{n} \sumt \sumtt \At_{\ell}\zt_{t''-1,\ell} \Rmc_{\Zt \ell}^{-1}(t''-1,t'-1)(\zt_{t'-1,\ell})\tran +o(\frac{1}{n}).
\end{align}
Then, we have
\begin{align}
    \rt_{t+1, \ell}^\ran &= \frac{1}{\sqrt{n}} \At_\ell \zt_{t\ell} - \frac{1}{\sqrt{n}} \sumt \sumtt \At_{\ell}\zt_{t''-1,\ell}\ \Rmc_{\Zt \ell}^{-1}(t''-1,t'-1) \left(\frac{1}{n}(\zt_{t'-1,\ell})\tran \zt_{t\ell}\right) +o(\frac{1}{n})\\
    &= \frac{1}{\sqrt{n}} \At_\ell \zt_{t\ell}- \frac{1}{\sqrt{n}} \sumtt  \At_{\ell}\zt_{t''-1,\ell}\ \sumj \sumt \Rmc_{\Zt \ell}^{-1}(t''-1,t'-1)\  \Rmc_{\Zt \ell} (t'-1, t-j)F_{t,j,\ell}+o(\frac{1}{n})\\
    &= \frac{1}{\sqrt{n}} \At_\ell (\zt_{t\ell} -\sum_{j=1}^{t} \zt_{t-j,\ell} F_{t,j,\ell})+o(\frac{1}{n})
\end{align}
Therefore, since $\At_{\ell}$ are i.i.d.\ Gaussian matrices, $\rt_{t+1,\ell}^\ran$ converges PL(2) to a Gaussian random variable $\Rt_{t+1,\ell}^\ran \sim \Norm(0, P_{t+1,\ell})$ such that,
\begin{align}
    P_{t+1,\ell} &= \Exp (\Zt_{t\ell} - \sum_{j=1}^{t}\Zt_{t-j,\ell} F_{t j\ell})\tran (\Zt_{t\ell} - \sum_{j=1}^{t}\Zt_{t-j,\ell} F_{t j\ell})
\end{align}
We can now write $\Rt_{t+1, \ell}$ as,
\begin{align}
    \Rt_{t+1,\ell} &= \Rt_{t+1,\ell}^\dete+\Rt_{t+1,\ell}^\ran\\
    &= \sum_{j=1}^t \Rt_{t-j+1,\ell} F_{t,j,\ell} + \Norm(0, P_{t+1,\ell}),
\end{align}
and by equation \eqref{eq:genvec} we have 
\begin{equation*}
    Q_{t+1} = \suml R_{t+1, \ell}, \qquad R_{t+1, \ell} = \Rt_{t+1, \ell}+ D_\ell \mu_{t\ell}.
\end{equation*}
This proves $\Mmc_{t}$ implies $\Mmc_{t+1}$.

\end{document}